\documentclass[12pt]{article}

\usepackage{graphicx}
\usepackage{hyperref}
\usepackage{amsmath,amsfonts,amssymb}
\usepackage{algorithmic,algorithm}
\usepackage[tight]{subfigure}

\newcommand{\appsection}[1]{\let\oldthesection\thesection
  \renewcommand{\thesection}{Appendix \oldthesection}
  \section{#1}\let\thesection\oldthesection}

\title{Polarimetric SAR Image Segmentation with B-Splines and a New Statistical Model}

\author{Alejandro C.\ Frery\and Julio Jacobo-Berlles\and   Juliana Gambini\and Marta E.\ Mejail%
\thanks{A.\ C.\ Frery is with the Instituto de Computa\c c\~ao, Universidade Federal de Alagoas, Brazil. M.\ E.\ Mejail, J. Gambini and J. Jacobo-Berlles are with the Departamento de Computaci\'on, Universidad de Buenos Aires, Argentina.}}

\begin{document}

\maketitle

\begin{abstract}
We present an approach for polarimetric Synthetic Aperture Radar (SAR) image region boundary detection based on the use of B-Spline active contours and a new model for polarimetric SAR data: the $\mathcal{G}_P^H$ distribution.
In order to detect the boundary of  a region, initial B-Spline curves are specified, either automatically or manually, and the proposed algorithm uses a deformable contours technique to find the boundary.
In doing this, the  parameters of the polarimetric $\mathcal{G}_P^H$ model for the data are estimated, in order to find the transition points between the region being segmented and the surrounding area.
This is a local algorithm since it works only on the region to be segmented.
Results of its performance are presented.
\end{abstract}

\section{Introduction}

Synthetic Aperture Radar (SAR) data have proven their importance in a number of applications, among which can be mentioned forestry, agriculture, analysis of geological and geomorphological features, thematic map updating, ocean oil spills, marine climatology, ice monitoring and deforestation (see, among others, \cite{Dierking2006,Dierking2002,ESA}).

Due to the active nature of the sensor, such images can be obtained at any time of the day, since the illumination source is carried by the sensing device.
SAR sensors work on the microwaves spectrum, so they are almost immune to adverse weather conditions and they are able to penetrate, to some extent, the surface of certain targets.
The first civilian SAR satellite was launched in 1978, and it was followed by a constellation of other similar sensors, mostly devoted to specific applications and in all cases operated at a single frequency and polarization.

The Shuttle Imaging Radar-C/X-band SAR (SIR-C/XSAR), launched in 1994, could be operated simultaneously at three frequencies, with two of them able to transmit and receive at both horizontal and vertical polarization.
This polarimetric capability provides a more complete description of the target~\cite{Zebker1991ProceedingsoftheIEEE}.

Polarimetric images are multiple complex-valued data sets requiring, thus, specialized models and algorithms.
Two main venues of research are followed for such description: electromagnetic modeling (see, for instance, \cite{SantAnna2008Sensors} and the references therein), and statistical laws.
We follow the second one.

Many techniques have been proposed for feature extraction and area classification in polarimetric SAR imagery.
Migliaccio et al.~\cite{Migliaccio2007} present a study on sea oil spill observation by means of polarimetric SAR data, based on the use of a constant false alarm rate filter.
Some techniques (see, for instance, \cite{schouDierking2000}) treat each channel individually and then fuse the results, but this approach does not exploit all the information these images convey.
Goudail et al.~\cite{Goudail2004} present a framework for designing algorithms that can solve detection, location and segmentation in polarimetric SAR images.
In these works the authors propose a definition of the contrast between regions with different polarimetric properties.
Horta et al.~\cite{Horta2008} perform polarimetric SAR image classification using the EM algorithm.

Conradsen et al.~\cite{Conradsen2003} model polarimetric SAR data with the complex Wishart distribution for edge detection.
Under that model, Davison et al.~\cite{Davidsonijrs} perform classification by maximum likelihood, while Ferro-Famil et al~\cite{Lee2004} present segmentation results.
Other distributions used for polarimetric SAR image segmentation are the $\mathcal{K}^P$ \cite{wang2008} and the $\mathcal{G}_P^0$ laws \cite{FreitasFreryCorreia:Environmetrics:03,Frery2007}.

Our approach is also based on the statistical description of the available information, which is in the form of a matrix in every pixel.
An edge is an ideal curve that divides two areas with different polarimetric scattering mechanisms, which yield different statistical properties in at least one of the available components; in our case, the roughness will be used as discriminatory feature.
This feature will be described as a real-valued parameter that indexes a new distribution for polarimetric SAR data, namely, the Harmonic $\mathcal G$ law (denoted $\mathcal G^H_P$).

To the knowledge of the authors of this paper, the literature shows no results of combining polarimetric  statistical properties with B-Spline based deformable contour methods.
Our method has the ability of operating on regions instead of over the whole image, which is a considerable virtue given the complexity and size of polarimetric SAR images.

The B-Spline approach has been widely used in curve representation for boundary detection~\cite{Zhang2008}, among other applications.
Contours formulated by means of B-Splines allow local control of the curve, have local representation, require few parameters and are intrinsically smooth.
A method for boundary estimation in noisy images based on B-Spline deformable contours, the Minimun Length Criterion and the Gaussian distribution is described in~\cite{Figueiredo2000}.
Gambini et al.~\cite{Gambiniijrs,GambMejJacFreryStatistic} developed techniques for boundary detection in univariate amplitude SAR imagery using the $\mathcal{G}_A^0$ distribution.

The technique proposed in this work is based on B-Spline boundary fitting, as proposed by Blake and Isard~\cite{Blake}, but tailored to the properties of polarimetric SAR imagery by means of the polarimetric $\mathcal{G}_P^H$  distribution as a general data model.
The polarimetric $\mathcal{G}_P^H$ model was recently developed, and presents an attractive choice for polarimetric SAR data segmentation.

This proposal for boundary extraction begins with the manual or automatic specification of initial regions of interest, determined by control points which generate a B-Spline curve.
Then, a series of radial segments are drawn on the image, and image data around them are extracted.
For each segment, the transition point, that is, the point belonging to the region boundary, is determined by parameter estimation from the data under the $\mathcal{G}_P^H$ model.
Then, for each region, the contour sought is given by the B-Spline curve that fits these transition points.

We apply this technique to simulated data in order to quantitatively assess its performance, and show its application to real polarimetric SAR data.
We also show that using roughness information from the three intensity components increases the discriminatory capability of the technique.

The structure of this paper is as follows.
Section~\ref{sec_pol} presents the new statistical model for polarimetric SAR data.
Section~\ref{boundary_detect} specifies the criterion used to determine the transition points and explains the region fitting algorithm, including the methodology employed for assessing the precision of the local edge detection technique.
Sections~\ref{sec_results} and~\ref{conclusiones} present the results of applying the technique to real SAR imagery and the conclusions, respectively.
Two appendices present details about single-channel SAR data (\ref{app:mono}) and about polarimetric laws, including algorithms for sampling from the random variables employed in our description (\ref{app:polarimetric}).

\section{Polarimetric SAR Data}\label{sec_pol}

Polarimetric SAR systems use antennae designed to transmit and receive electromagnetic waves of a specific polarization, being the two most common ones the horizontal linear or H, and vertical linear or V.
Due to the possible change in polarization of the scattered wave, radar antennae are designed to receive the different polarization components simultaneously and, therefore, HH, VV, HV and VH data will be available in a full polarimetric system.
HV and VH channels are identical in an ideal and perfectly calibrated monostatic radar system, so one of them will be discarded in the following.

Then, we define the complex random vector $\mathbf{Z}$ as
\begin{equation}
\mathbf{Z}=[ Z_{HH},Z_{HV},Z_{VV}] ^{t},
\label{return}
\end{equation}
where `$t$' denotes transposition and $Z_{HH}$, $Z_{HV}$ and $Z_{VV}$ denote the corresponding components of the backscattered electromagnetic fields, with the first subscript indicating the polarization of the transmitted electromagnetic field and the second subscript indicating the polarization of the detected component of the backscattered electromagnetic field.


Multi-look signal processing is frequently applied in order to enhance the signal-to-noise ratio.
We define the multi-look complex matrix $\mathbf{Z}^{(n) }$ of size $3\times 3$ as
\begin{equation}
\mathbf{Z}^{(n) }=\frac{1}{n}\sum\limits_{k=1}^{n}\mathbf{Z}(k) \mathbf{Z}^{\ast t}(k),
\label{ec_Z_multi-look}
\end{equation}%
where `$*$' denotes the complex conjugate, $n$ is the number of looks  and $\mathbf{Z}(k) $, $1 \leq k \leq n$, are random vectors of the form defined in~\eqref{return}, so each term of summation in~\eqref{ec_Z_multi-look} is given by
$$
\mathbf{Z}( k) \mathbf{Z}^{\ast t}(k)=\left[
\begin{array}
[c]{ccc}%
Z_{HH}Z_{HH}^* & Z_{HH}Z_{HV}^* & Z_{HH}Z_{VV}^*\\
Z_{HV}Z_{HH}^* & Z_{HV}Z_{HV}^* & Z_{HV}Z_{VV}^*\\
Z_{VV}Z_{HH}^* & Z_{VV}Z_{HV}^* & Z_{VV}Z_{VV}^*
\end{array}
\right].
$$
In this Hermitian matrix the diagonal elements are real numbers, while the off-diagonal elements are complex numbers with non-null imaginary components.

We will follow the multiplicative paradigm,  so the returned polarimetric data will be considered to be the result of the product between backscatter variability and polarimetric speckle, given by
\begin{equation}
\left[
\begin{array}{c}
Z_{HH} \\
Z_{HV} \\
Z_{VV}%
\end{array}%
\right] =\sqrt{X}\left[
\begin{array}{c}
Y_{HH} \\
Y_{HV} \\
Y_{VV}%
\end{array}%
\right] ,  \label{ec_modelo_mult}
\end{equation}%
where $\mathbf{Z}=\left[ Z_{HH},Z_{HV},Z_{VV}\right] ^{t}$ and $\mathbf{Y}=\left[
Y_{HH},Y_{HV},Y_{VV}\right] ^{t}$ are complex random vectors~\cite{LeePottier2009PolarimetricRadarImaging}.
The random variable $X$ is scalar, has unitary mean, and models backscatter variability due to the heterogeneity of the sensed area.
The random vector $\mathbf{Y}$ represents the polarimetric speckle noise and the mean values of each of its components determine the mean values of each of the corresponding components of $\mathbf{Z}$.

The multi-look polarimetric speckle matrix $\mathbf{Y}^{( n) }$ is defined as
\begin{equation}
\mathbf{Y}^{( n) }=\frac{1}{n}\sum\limits_{k=1}^{n}\mathbf{Y}( k) \mathbf{Y}^{\ast t}( k) . \label{ec_Y_multi-look}
\end{equation}%
Then, from~\eqref{ec_Z_multi-look}, \eqref{ec_modelo_mult} and \eqref{ec_Y_multi-look} one has
that $\mathbf{Z}^{( n) }=X\mathbf{Y}^{( n) }$, so equation~\eqref{ec_modelo_mult} can be rewritten as
$$
\mathbf{Z}^{( n) }=\frac{X}{n}\sum\limits_{k=1}^{n}\mathbf{Y}( k) \mathbf{Y}^{\ast t}( k),
\label{ec_MM_multi-look_expand}
$$
where
$$
\mathbf{Y}( k) \mathbf{Y}^{\ast t}( k) =
\left[
\begin{array}{ccc}
Y_{HH}Y_{HH}^* & Y_{HH}Y_{HV}^* & Y_{HH}Y_{VV}^*\\
Y_{HV}Y_{HH}^* & Y_{HV}Y_{HV}^* & Y_{HV}Y_{VV}^*\\
Y_{VV}Y_{HH}^* & Y_{VV}Y_{HV}^* & Y_{VV}Y_{VV}^*
\end{array}%
\right].
$$

\subsection{A new polarimetric distribution}\label{sec_gh_pol}

If we consider that the components of $\mathbf{Y}( k) $ exhibit a Multivariate Complex Gaussian distribution, then $n\mathbf{Y}^{( n) }$ will have a Centered Complex Wishart distribution, as presented in \ref{app:polarimetric}.
So the density function of $\mathbf{Y}^{( n) }$ is given by
\begin{equation}
f_{\mathbf{Y}^{( n) }}( \mathbf{y}) =\frac{
n^{3n}\left| \mathbf{y}\right| ^{n-3}\exp (
-n\mathrm{Tr}( \Sigma_{\mathbf Y}^{-1}\mathbf{y}) )}{\pi ^{3}\Gamma ( n) \cdots
\Gamma ( n-2) \left| \Sigma _{\mathbf{Y}}\right| ^{n}}
 ,
\label{ec_dens_Y}
\end{equation}
for $n\geq 3$ and for $\mathbf Y\in \mathbb{C}^{3\times 3}$, where $\mathrm{Tr}$ is
the trace, $| \cdot | $ denotes the determinant of a
matrix and $\Sigma_{\mathbf Y}$ is the covariance matrix of $\mathbf Y$.

In order to find the density function of $\mathbf{Z}^{( n) }$, the
following integral has to be computed:
\begin{equation}
f_{\mathbf{Z}^{( n) }}( \mathbf{z}) =\int\nolimits_{%
\mathbb{R}_{+}}f_{\mathbf{Z}^{( n) }\mid X=x}( \mathbf{z}%
) \,f_{X}( x) \,dx.  \label{ec_integral_dens_Z}
\end{equation}%
Using~(\ref{ec_dens_Y}) we find that
\begin{equation}
f_{\mathbf{Z}^{( n) }\mid X=x}( \mathbf{z})
=x^{-3^{2}}f_{\mathbf{Y}^{( n) }}( x^{-1}\mathbf{z}) .
\label{ec_dens_Z_condicional}
\end{equation}%
We consider that the variability of backscatter, modeled by random variable $X$, follows an Inverse Gaussian distribution~\cite{Seshadri93} with unitary mean $X \sim IG(\omega,1)$ (see \ref{app:polarimetric}) with density function
\begin{equation}
f_{X}( x) =\sqrt{\frac{\omega }{2\pi x^{3}}}\exp \Bigl\{ -\frac{\omega}{2} \frac{( x-1) ^{2}}{x}\Bigr\},  \label{ec_dens_gauss_inv}
\end{equation}%
where $x>0$ and $\omega>0$ is the roughness parameter.
This situation, denoted $X\sim IG(\omega,1)$, is a particular case of the Generalized Inverse Gaussian distribution~\cite{Jorgensen82}, whose use for backscatter modeling was proposed by Frery et al.~\cite{frery97}.

Now, from~(\ref{ec_integral_dens_Z}), (\ref{ec_dens_Z_condicional}) and~(\ref{ec_dens_gauss_inv}) we obtain
$$f_{\mathbf{Z}^{( n) }}( \mathbf{z}) =\frac{\sqrt{\frac{%
2}{n}}n^{3n}e^{\omega }\omega ^{3n+1}\left| \mathbf{z}\right| ^{n-3}}{\pi
^{3}\Gamma ( n) \cdots \Gamma ( n-2) \left| \Sigma_{\mathbf Y}\right| ^{n}}
\frac{K_{3n+1/2}( \sqrt{\omega ( 2n\mathrm{Tr}(\Sigma_{\mathbf Y}^{-1}\mathbf{z}) +\omega ) }) }{( \omega ( 2n%
\mathrm{Tr}( \Sigma_{\mathbf Y}^{-1}\mathbf{z}) +\omega ) )
^{\frac{3}{2}n+\frac{1}{4}}}.  \label{ec_dens_Z}
$$
We denote this situation $\mathbf Z^{(n)}\sim\mathcal G_P^H(\omega,\Sigma_{\mathbf Y})$: the Harmonic Polarimetric distribution.
\ref{app:mono} presents the intensity channel version of this and other related laws, while \ref{app:polarimetric} provides a complete account of the polarimetric laws derived from the multiplicative model and the Generalized Inverse Gaussian distribution for the backscatter.

The Bessel function $K_{3n+1/2}$ above can be computed using a closed formula:
$$
K_{np+1/2}(\nu)=\sqrt{\frac{\pi}{2\nu}}e^\nu\sum_{k=0}^{np}\frac{(np+k)!}{k!(np-k)(2\nu)^k},\label{formKBesselFrac}
$$
with
$$
\nu=\sqrt{\omega(  2n\mathrm{Tr}(
\Sigma_{\mathbf{Y}}^{-1}\mathbf{z})  +\omega) },
$$
alleviating, thus, the numerical issues that the evaluation of this function imposes in the general case.

\subsection{Parameter Estimation}\label{estima_parametros}

The estimation of the roughness parameter $\omega$ can be done using the first and second order moments of the diagonal elements of $\mathbf{Z}^{(n) }$.
The components of the main diagonal of $\mathbf{Z}^{(n)}$ are given by
\[
Z_{i,i}^{(n)}=\frac{X}{n}\sum_{k=1}^{n}\left|
Y_{k,i}\right|  ^{2},\text{ with }i\in\left\{
HH,HV,VV\right\}  ,
\]
where the random variables $X$ and ${n}^{-1}\sum_{k=1}^{n}\left|
Y_{k,i}\right|  ^{2}$ are such that $X\sim IG(  \omega,1)$ and ${n}^{-1}\sum_{k=1}^{n}\left|
Y_{k,i}\right|  ^{2} \sim \sigma^{2}_i\Gamma(n,2n)$, where $\sigma^{2}_i$ is the mean of $Z_{i,i}^{(n)}$. This is equivalent to considering $Z_{i,i}^{(n)}$\ as the result of the product of a $IG(  \omega,\sigma_{i}^{2})$ distributed random variable and a
 $\Gamma(  n,2n)  $ distributed random variable, because a
 $\sigma_{i}^{2}IG(  \omega,1)  $ distributed random variable
is  $IG(  \omega,\sigma_{i}^{2})  $ distributed.
This, implies that $Z_{i,i}^{(n)}$ is a $\mathcal{G}_{I}^{H}(  \omega,\sigma_{i}^{2},n)  $ distributed random variable (see \ref{app:mono}).
Thus, we can estimate these parameters as in the case of the univariate intensity data.

Then, the $r$th-order moment of the return $Z$ is
\begin{equation*}
\mathbb{E}[(Z_{i,i}^{(n)})^r]  = \Bigl(\frac{\eta}{n}\Bigr)^r e^{\omega} \sqrt{\frac{2\omega}{\pi}}
K_{r-\frac{1}{2}}(\omega)\frac{\Gamma(n+r)}{\Gamma(n)}.
\label{momGI_del_retorno}
\end{equation*}
Now calling $ m_{1i}= \widehat{\mathbb{E}}[Z_{i,i}^{(n)}]$ and $ m_{2i}= \widehat{\mathbb{E}}[(Z_{i,i}^{(n)})^2]$, estimates of $\omega$ are given by
\begin{equation*}
\widehat{\omega}_i =\frac{1}{\frac{n}{n+1}\frac{ m_{2i}}{m_{1i}^{2}}-1},
\label{ec_omega_est_i}
\end{equation*}
and the estimates of $\sigma_{i}^{2}$ are given by
\begin{equation}
\widehat{\sigma_{i}^{2}}=m_{1i}\label{SigmaEstimado}%
\end{equation}
for $i \in \{ HH, HV, VV \}$.

The value of the parameter $\omega$ common to the three components will be chosen as the one that minimizes the total error $\epsilon$ given by
\begin{equation*}
\epsilon=\sum_{i} \sum_{k_i} \left(f_{Z_i}\left(\omega,\widehat{\sigma}_i; z_{k_i} \right) - h\left(z_{k_i} \right)\right)^2,
\end{equation*}
with $i \in \{ HH, HV, VV \}$, $k_i \in \{1, \ldots ,N \}$, where $N$ is the sample size, $Z_i \sim \mathcal{G}_I^H\left(\omega, \sigma_i^2, n \right)$ and $h$ is a histogram.
The parameters that compose the correlation matrix are given in table~\ref{los_parametros}.

\begin{table}[hbt]
\centering
\caption{Statistical parameters of the correlation matrix.}\label{los_parametros}
\begin{tabular}[c]{|l|c|c|c|}\hline
 $\  $ &HH  & HV& VV\\\hline
HH & $\sigma_{HH}^2$ & $a_{HHHV}+jb_{HHHV}$ & $a_{HHVV}+jb_{HHVV}$ \\
HV&$\ $ & $\sigma_{HV}^2$ & $a_{HVVV}+jb_{HVVV}$ \\
VV&$\ $& $\ $ & $\sigma_{VV}^2$ \\\hline
\end{tabular}
\end{table}

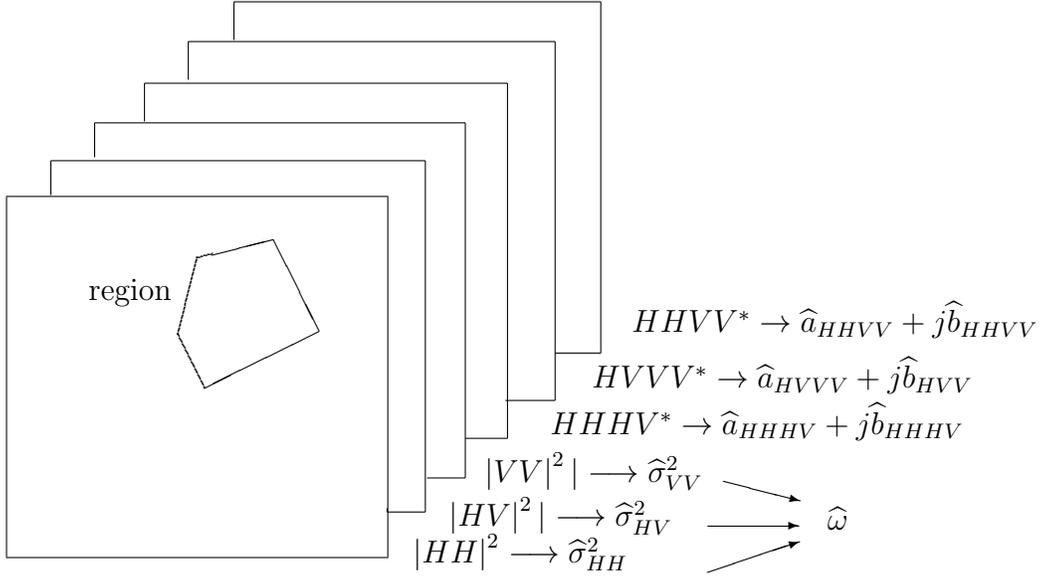
\begin{figure}
 \centering
 \setlength{\unitlength}{3947sp}%
\begingroup\makeatletter\ifx\SetFigFont\undefined%
\gdef\SetFigFont#1#2#3#4#5{%
  \reset@font\fontsize{#1}{#2pt}%
  \fontfamily{#3}\fontseries{#4}\fontshape{#5}%
  \selectfont}%
\fi\endgroup%
\begin{picture}(6412,3514)(1289,-3186)
\thinlines
\put(1301,-3174){\framebox(2388,2263){}}
\put(1576,-899){\line( 0, 1){216}}
\put(1576,-683){\line( 1, 0){2352}}
\put(3928,-683){\line( 0,-1){2208}}
\put(3928,-2891){\line(-1, 0){240}}
\put(3688,-2891){\line( 0, 1){ 24}}
\put(1851,-661){\line( 0, 1){216}}
\put(1851,-445){\line( 1, 0){2328}}
\put(4179,-445){\line( 0,-1){2232}}
\put(4179,-2677){\line(-1, 0){240}}
\put(3939,-2677){\line( 0, 1){  0}}
\put(2164,-436){\line( 0, 1){  0}}
\put(2164,-436){\line( 0, 1){240}}
\put(2164,-196){\line( 1, 0){2280}}
\put(4444,-196){\line( 0,-1){2232}}
\put(4444,-2428){\line(-1, 0){264}}
\put(4180,-2428){\line( 0, 1){  0}}
\put(2439,-174){\line( 0, 1){240}}
\put(2439, 66){\line( 1, 0){2304}}
\put(4743, 66){\line( 0,-1){2256}}
\put(4743,-2190){\line(-1, 0){312}}
\put(4431,-2190){\line( 0, 1){  0}}
\put(2726, 76){\line( 0, 1){240}}
\put(2726,316){\line( 1, 0){2304}}
\put(5030,316){\line( 0,-1){2208}}
\put(5030,-1892){\line(-1, 0){288}}
\put(4742,-1892){\line( 0, 1){  0}}
\put(5700,-3270){\vector( 3, 1){580}}
\put(5800,-2700){\vector( 4,-1){484}}
\put(5700,-2980){\vector( 1, 0){576}}
\put(2589,-1274){\line( 4, 1){384}}
\put(2973,-1178){\line( 1,-2){288}}
\put(3261,-1754){\line(-2,-1){720}}
\multiput(2541,-2114)(-3.73333,7.46667){46}{\makebox(1.6667,11.6667){\SetFigFont{5}{6}{\rmdefault}{\mddefault}{\updefault}.}}
\multiput(2373,-1778)(2.03390,8.13559){60}{\makebox(1.6667,11.6667){\SetFigFont{5}{6}{\rmdefault}{\mddefault}{\updefault}.}}
\multiput(2493,-1298)(8.72727,2.18182){12}{\makebox(1.6667,11.6667){\SetFigFont{5}{6}{\rmdefault}{\mddefault}{\updefault}.}}
\put(3851,-3211){\makebox(0,0)[lb]{\smash{{\SetFigFont{12}{14.4}{\rmdefault}{\mddefault}{\updefault}$\left|HH\right|^2 \longrightarrow \widehat{\sigma}^2_{HH}$}}}}
\put(4076,-2980){\makebox(0,0)[lb]{\smash{{\SetFigFont{12}{14.4}{\rmdefault}{\mddefault}{\updefault}$\left|HV\right|^2|\longrightarrow \widehat{\sigma}^2_{HV}$}}}}
\put(4301,-2700){\makebox(0,0)[lb]{\smash{{\SetFigFont{12}{14.4}{\rmdefault}{\mddefault}{\updefault}$\left|VV\right|^2|\longrightarrow \widehat{\sigma}^2_{VV}$}}}}
\put(4714,-2400){\makebox(0,0)[lb]{\smash{{\SetFigFont{12}{14.4}{\rmdefault}{\mddefault}{\updefault}$HHHV^* \rightarrow \widehat{a}_{HHHV}+j \widehat{b}_{HHHV}$}}}}
\put(4976,-2111){\makebox(0,0)[lb]{\smash{{\SetFigFont{12}{14.4}{\rmdefault}{\mddefault}{\updefault}$HVVV^* \rightarrow \widehat{a}_{HVVV}+j \widehat{b}_{HVV}$}}}}
\put(5226,-1761){\makebox(0,0)[lb]{\smash{{\SetFigFont{12}{14.4}{\rmdefault}{\mddefault}{\updefault}$HHVV^* \rightarrow \widehat{a}_{HHVV}+j \widehat{b}_{HHVV}$}}}}
\put(1814,-1561){\makebox(0,0)[lb]{\smash{{\SetFigFont{12}{14.4}{\rmdefault}{\mddefault}{\updefault}region}}}}
\put(6455,-3000){\makebox(0,0)[lb]{\smash{{\SetFigFont{12}{14.4}{\rmdefault}{\mddefault}{\updefault}$\widehat{\omega}$}}}}
\end{picture}%
 \caption{Example of covariance matrix and $\omega$ parameter estimation for a particular region.}
 \label{fig:bandas}
\end{figure}

Figure~\ref{fig:bandas} shows the covariance matrix and the images necessary to estimate the statistical parameters of the covariance matrix for a particular region, where the arrows relate the images to the corresponding estimated parameters. As the covariance matrix is Hermitian, only the upper triangle and the diagonal are displayed.

The off-diagonal elements $Z_{i,\ell}^{(n)}$ in $\mathbf{Z}^{(n)}%
$ are given by
\[
Z_{i,\ell}^{(n)}=\frac{1}{n}\sum_{k=1}^{n}Z_{k,i}%
Z_{k,\ell}^{\ast}=X \frac{1}{n}\sum_{k=1}^{n}Y%
_{k,i}Y_{k,\ell}^{\ast}\text{, \textrm{for }}i\neq \ell,\text{ and }%
i,\ell\in\left\{  HH, HV, VV\right\}  .
\]
Due to the independence between $X$ and $Y_{i}$, with $i\in\left\{  HH, HV, VV\right\}  $, and taking into account that $\mathbb{E}\left[X\right]  =1$, we have that
\[
\mathbb{E}\left[  Z_{i,\ell}^{(n)}\right]  =\mathbb{E}\left[  Z_{i}Z%
_{\ell}^{\ast}\right]  =\mathbb{E}\left[  X\right]  \mathbb{E}\left[  Y%
_{i}Y_{\ell}^{\ast}\right]  =\mathbb{E}\left[  Y_{i}Y_{\ell}%
^{\ast}\right]  .%
\]
As $\mathbb{E}\left[  Y_{i}Y_{\ell}^{\ast}\right]  =\left(
a_{i\ell}+jb_{i\ell}\right)  \sigma_{i}\sigma_{\ell}$, then $\widehat{a}_{i\ell}$ and $\widehat{b}_{i\ell}$ can be estimated as
\[
\widehat{a}_{i\ell}+j\widehat{b}_{i\ell}=\frac{\widehat{\mathbb E}\left[  Z%
_{i}Z_{\ell}^{\ast}\right]  }{\widehat{\sigma}_{i}\widehat{\sigma}_{\ell}%
},%
\]
where $\widehat{\mathbb{E}}\left[  Z_{i}Z_{\ell}^{\ast}\right]$ and $\widehat{\sigma}_{i}$ are estimated from the observed data $\mathbf{Z}=[Z_{HH},Z_{HV},Z_{VV}]^{t}$ and using equation~(\ref{SigmaEstimado}), respectively.

\subsection{Parameter interpretation}\label{parestim_sec}

One of the most important features of the $\mathcal{G}^{H}$ (both intensity and polarimetric) distribution is that the estimated values of the parameter $\omega$ have immediate interpretation in terms of roughness.
For values of $\omega$ near zero, the imaged area presents very heterogeneous gray values, as is the case of urban areas in polarimetric SAR images.
As we move to less heterogeneous areas like forests, the value of $\omega$ grows, reaching its highest values for homogeneous areas like pastures and certain types of crops.
This is the reason why this parameter is regarded to as a roughness or texture measure.

The parameters $\sigma_{i}^{2}$, with $i \in \{ HH, HV, VV \}$, are the average intensities for each polarization.
The parameters $a_{k\ell} + j \ b_{k\ell} $, with $k,\ell \in \{ HH, HV, VV \}$, are the correlation coefficients among the three polarimetric images.

\begin{figure}[htb]
	\centering
		\includegraphics[width=.7\linewidth]{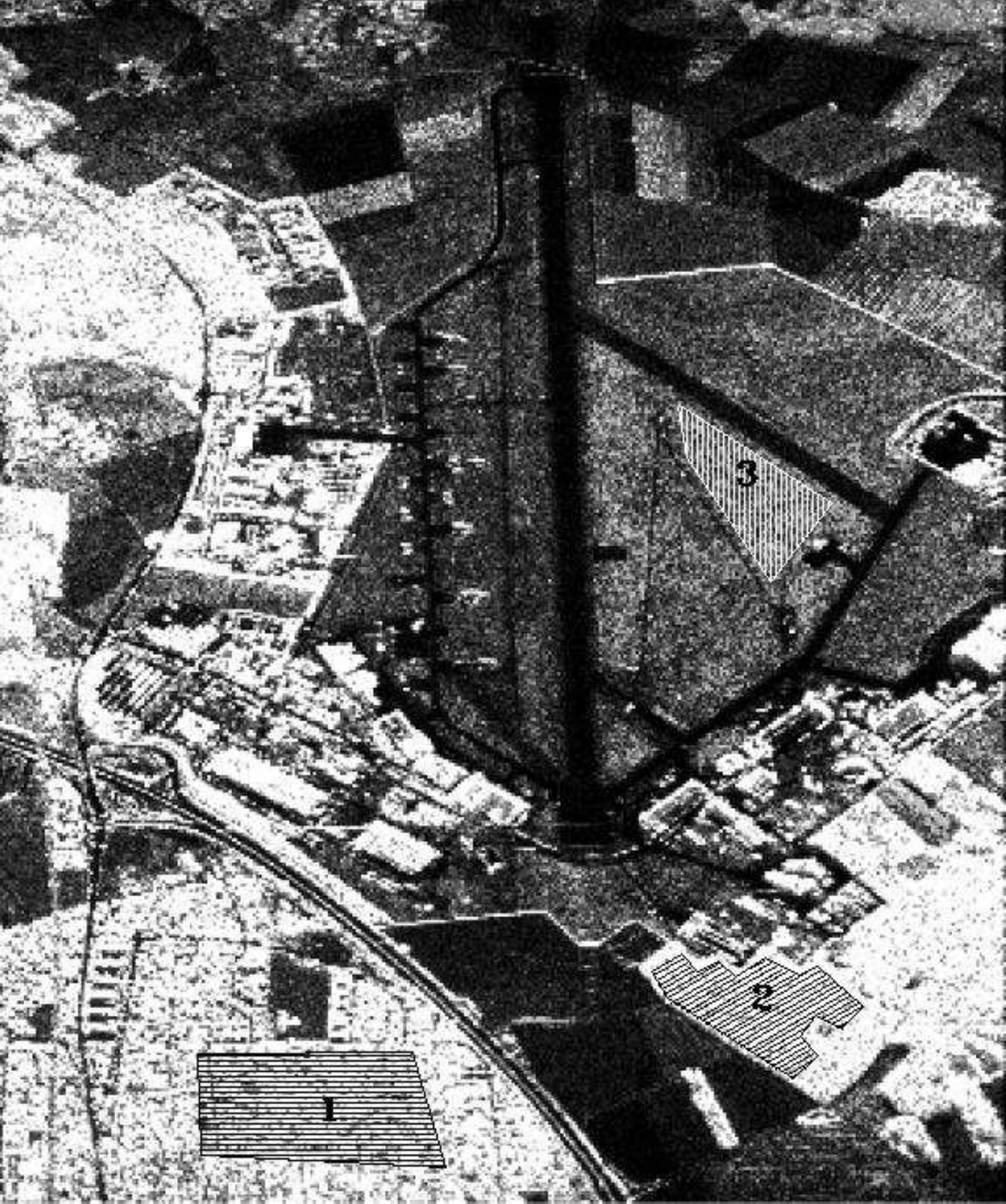}
	\caption{L-band E-SAR image from a zone near the city of We\ss ling, Bayern, Germany with three regions of interest: Urban~(1), Forest~(2) and Pasture~(3)}
	\label{fig:ZhvZvvZhh3looks_wrois}
\end{figure}

In order to check the capability of the $\mathcal G_P^H$ model for describing polarimetric data, an E-SAR image of We\ss ling (Bayern, Germany) was used~\cite{Horn1998}.
This single look image, shown in Figure~\ref{fig:ZhvZvvZhh3looks_wrois}, was obtained in L band, and it exhibits an airport, urban areas, forest and pastures.
From this one look complex polarimetric image, three intensity images were generated by taking one of every three columns (azimuth direction).
These images were then averaged yielding a three looks intensity image.
The parameters were estimated using samples from the last three targets, which are shown in Figure~\ref{fig:ZhvZvvZhh3looks_wrois}.

The estimated covariance matrices for urban, forest and pasture areas are given in~\eqref{matriz_sigmay_ciudad}, \eqref{matriz_sigmay_bosque} and~\eqref{matriz_sigmay_pasto}, respectively:
\begin{align}
\widehat\Sigma_{\text{u}}=&\left[
\begin{array}[c]{ccc}%
962892	&	19171-j3579  &	-154638+j191388\\
&	56707	&	-5798+j16812\\
& & 472251\\
\end{array}
\right], \label{matriz_sigmay_ciudad}\\%
\widehat\Sigma_{\text{f}}=&\left[
\begin{array}
[c]{ccc}%
360932 & 11050+j3759 & 63896+j1581\\
& 98960 &  6593+j6868\\
& &  208843\\
\end{array}
\right], \label{matriz_sigmay_bosque}\\%
\widehat	\Sigma_{\text{p}}=&\left[
\begin{array}
[c]{ccc}%
32556 &	556+j787 & 24046-j27287\\
& 1647	& -146-j482\\
& & 61028\\
\end{array}
\right]. \label{matriz_sigmay_pasto}
\end{align}

Table \ref{tabla__omega} shows the estimated values of $\omega$ for urban, forest and pasture targets, respectively, in $HH$, $HV$ and $VV$ polarizations.
These values are comparable in different frequencies when estimated in areas with similar textures.
Recall from equation~\eqref{ec_dens_gauss_inv} that $\omega$ is only allowed positive values; c.f.\ \ref{app:polarimetric} for more details.

\begin{table}[htb]
\begin{center}
\caption{Estimated roughness parameters}\label{tabla__omega}
\renewcommand{\arraystretch}{1.3}
\begin{tabular}{|c|c|c|c|c|} \hline
$\widehat\omega$ & \bfseries HH        & \bfseries HV         & \bfseries VV         & \bfseries Average       \\ \hline
\bfseries Urban &  0.17   &  0.94  &   0.19  &   0.43  \\ \hline
\bfseries Forest & 10.22   &  8.53  &   10.55  &   9.77 \\ \hline
\bfseries Pasture & 19.88   &  22.54  &   18.32  &   20.24 \\ \hline
\end{tabular}
\end{center}
\end{table}

It is noteworthy that the estimated values of the parameters that index any polarimetric distribution derived from the multiplicative model do not depend, in principle, from the number of looks; the roughness of a target is preserved, regardless that measure of the signal-to-noise ratio.

\begin{figure}[htb]
\centering
\subfigure[HH $\mathcal{G}_{P}^{H}$]{\includegraphics[width=.32\linewidth]{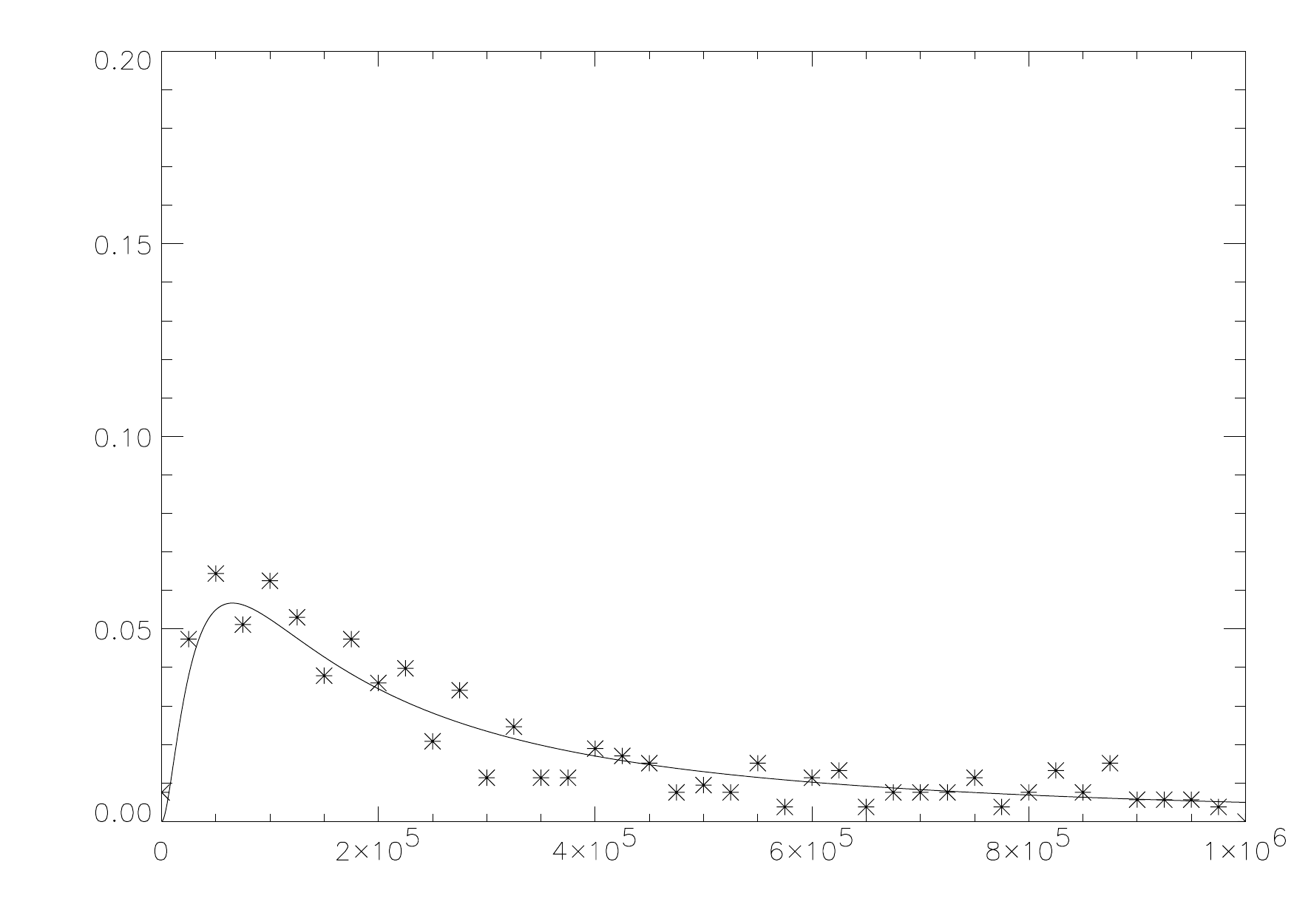}\label{fig_city_hh_gh}}
\subfigure[HV $\mathcal{G}_{P}^{H}$]{\includegraphics[width=.32\linewidth]{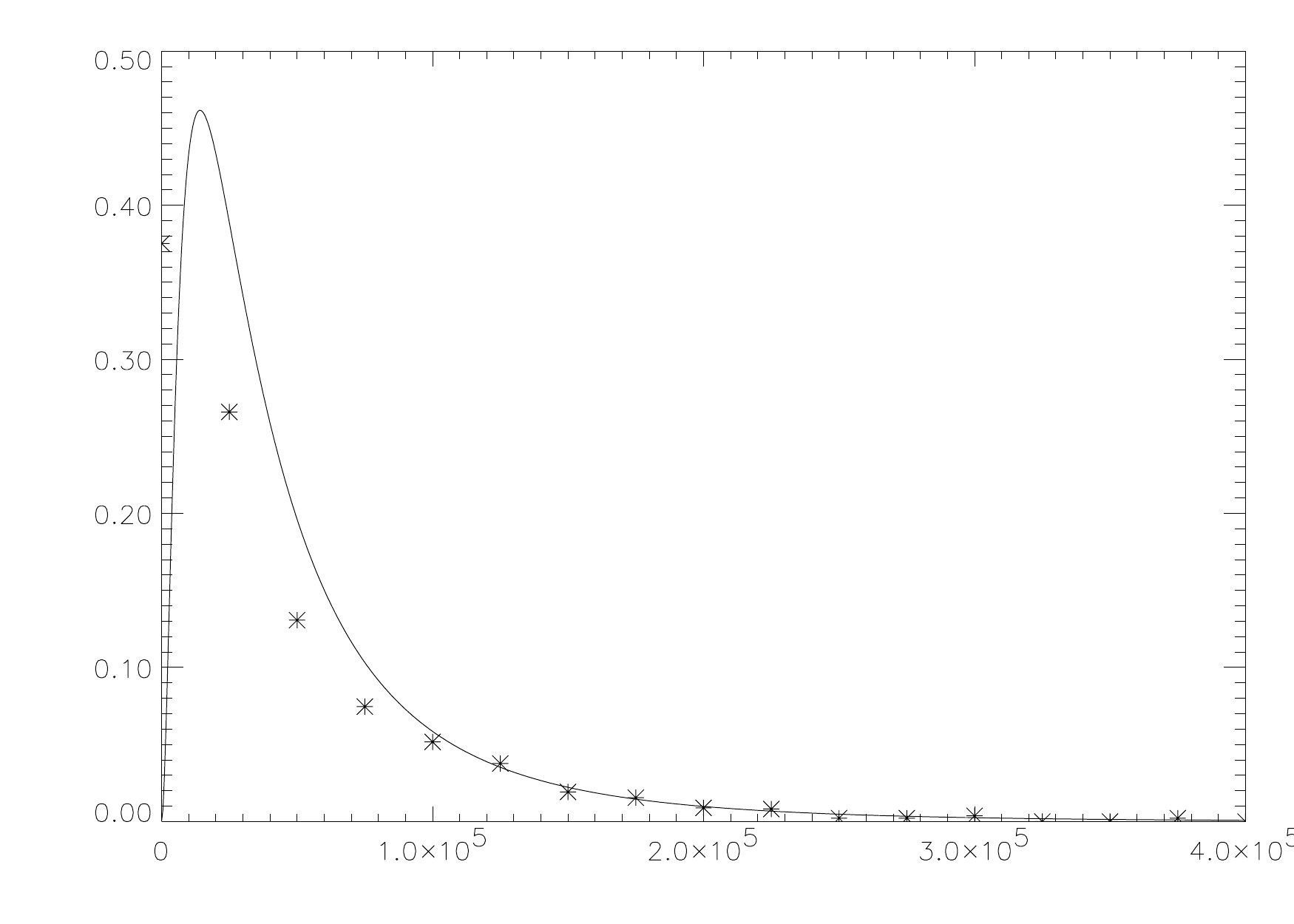}\label{fig_city_hv_gh}}
\subfigure[VV $\mathcal{G}_{P}^{H}$]{\includegraphics[width=.32\linewidth]{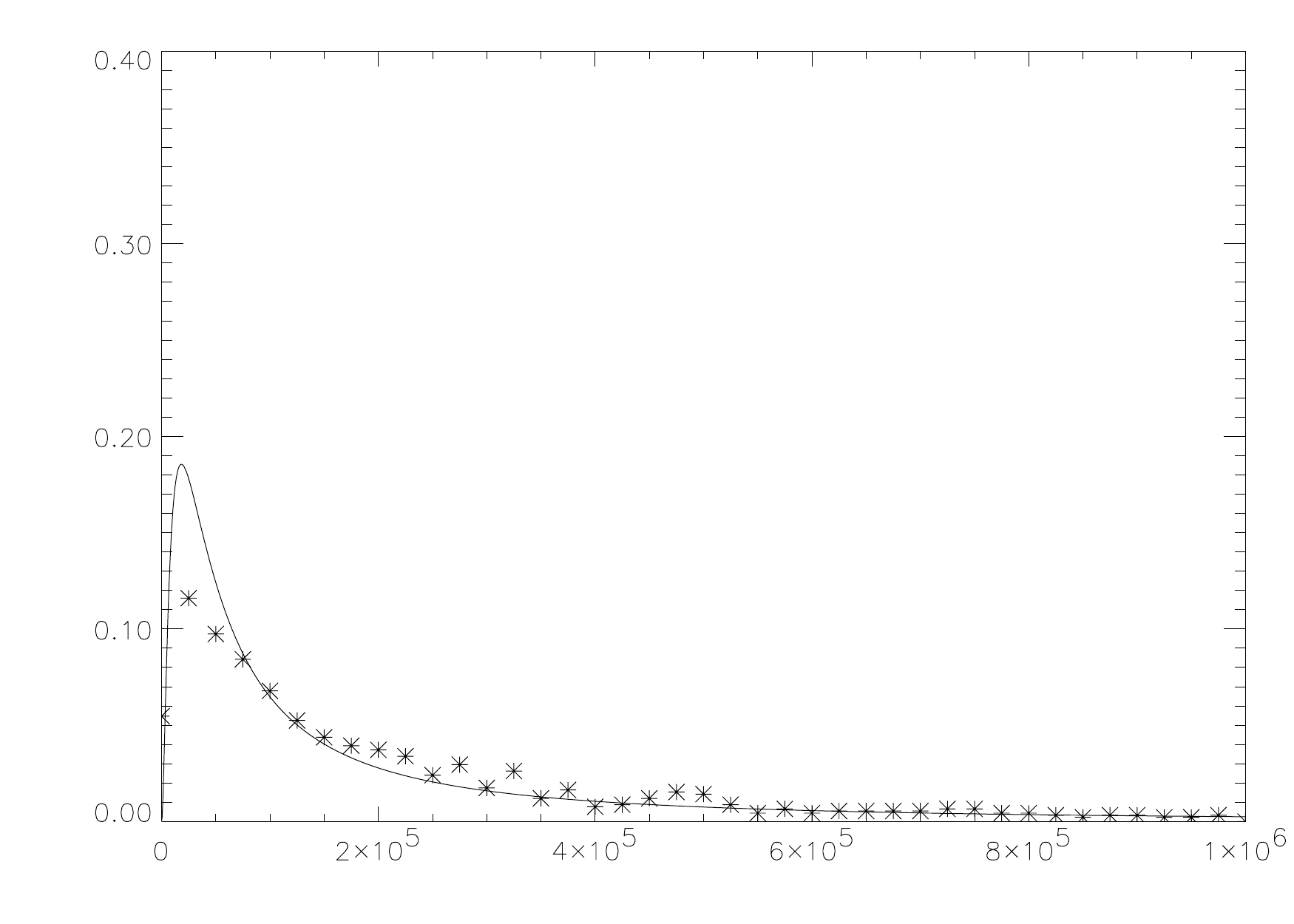}\label{fig_city_v_gh}}
\caption{Histograms and estimated densities for urban data}\label{fig_ajustes_city}
\end{figure}

\begin{figure}[htb]
\centering
\subfigure[HH $\mathcal{G}_{P}^{H}$]{\includegraphics[width=.32\linewidth]{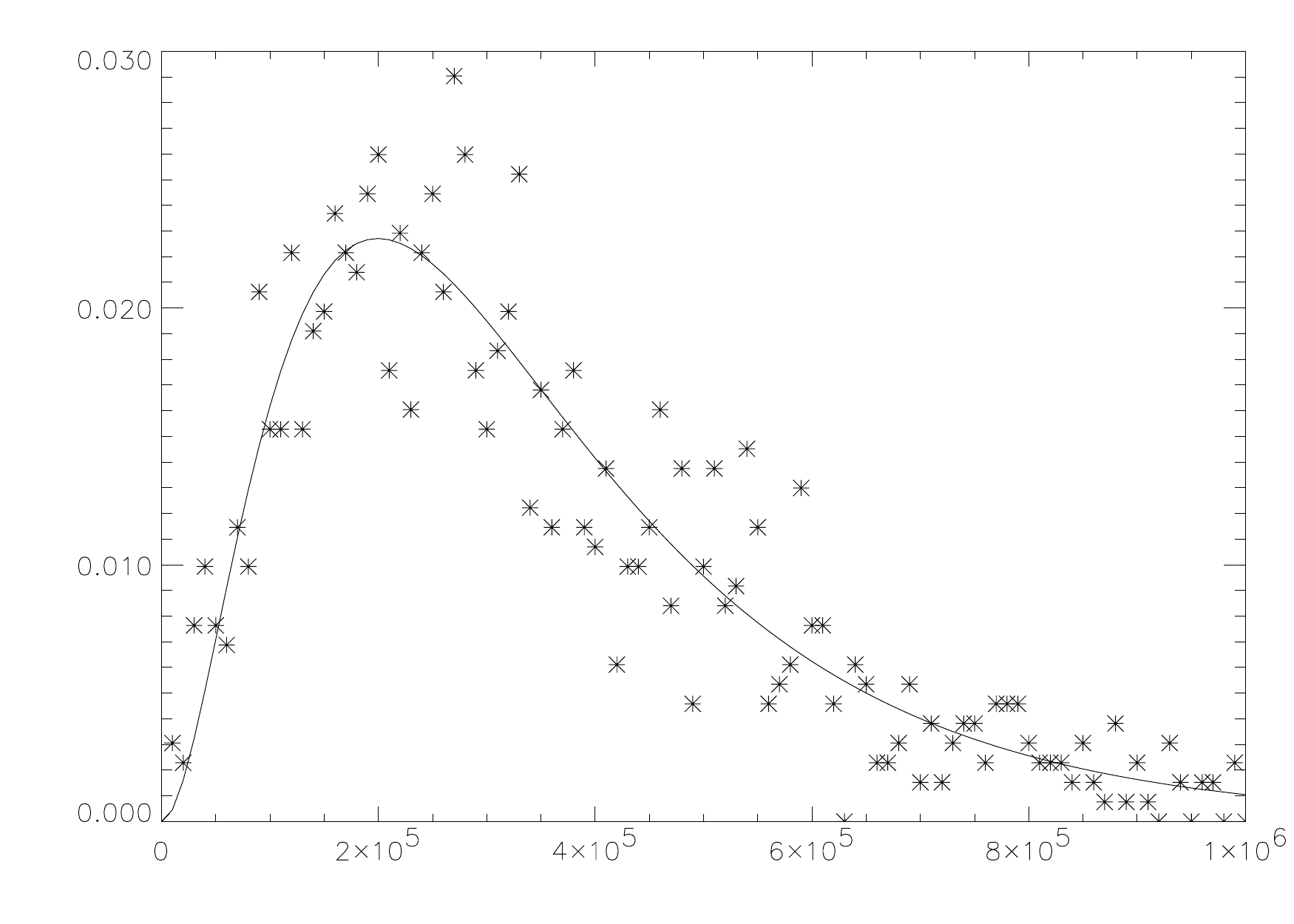}\label{fig_forest_hh_gh}}
\subfigure[HV $\mathcal{G}_{P}^{H}$]{\includegraphics[width=.32\linewidth]{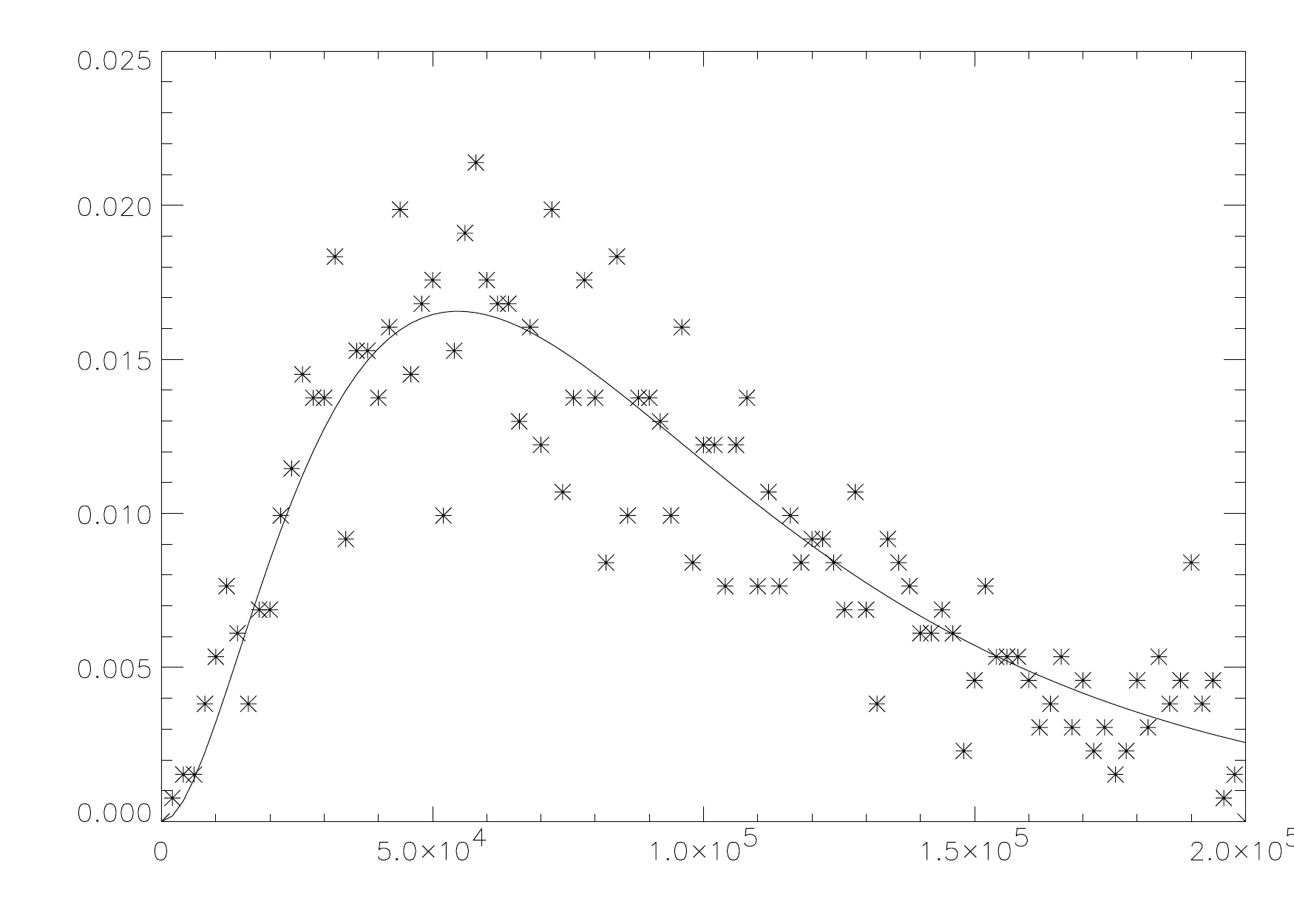}\label{fig_forest_hv_gh}}
\subfigure[VV $\mathcal{G}_{P}^{H}$]{\includegraphics[width=.32\linewidth]{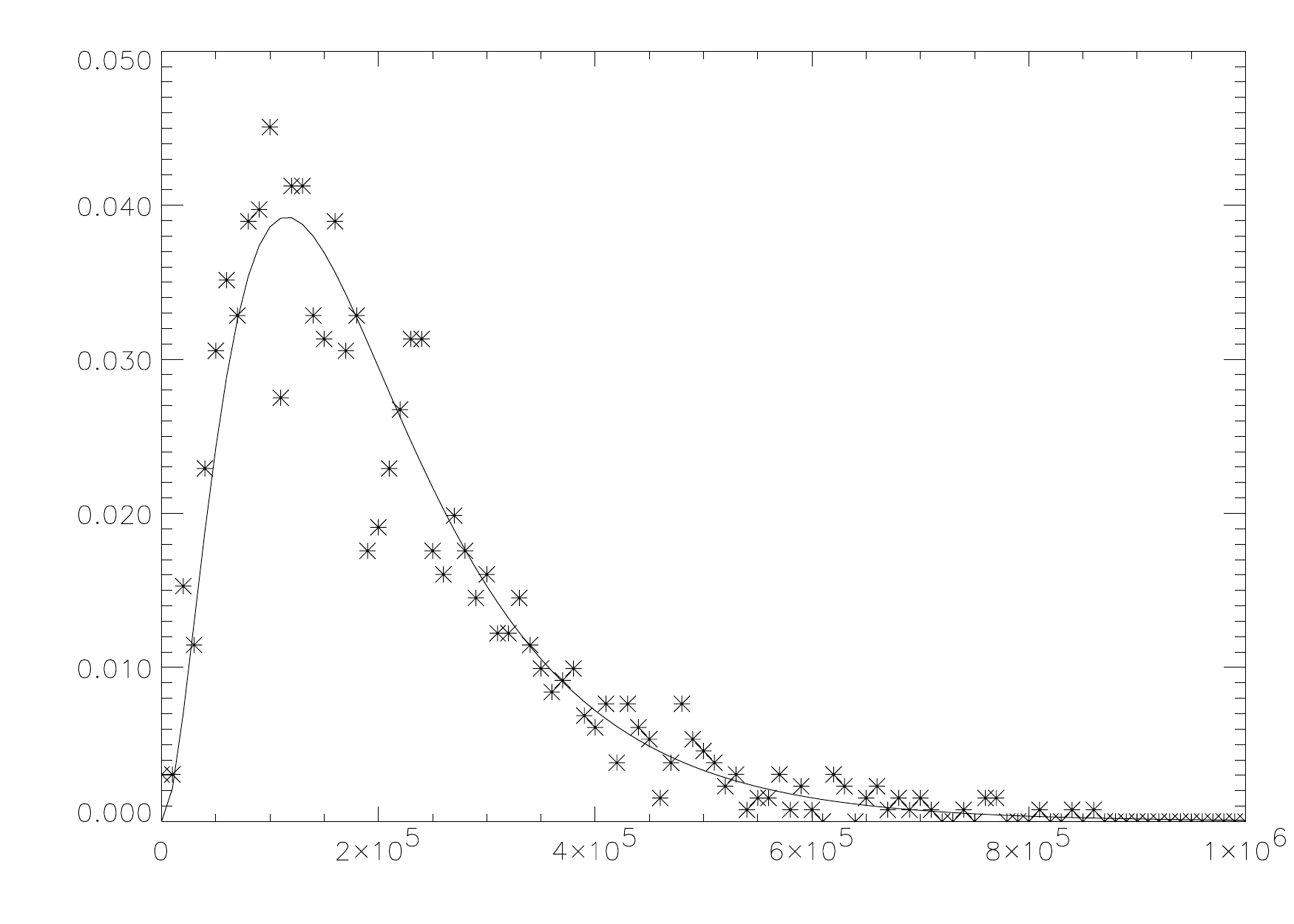}\label{fig_forest_v_gh}}
\caption{Histograms and estimated densities for forest data}\label{fig_ajustes_forest}
\end{figure}

\begin{figure}[htb]
\centering
\subfigure[HH $\mathcal{G}_{P}^{H}$]{\includegraphics[width=.32\linewidth]{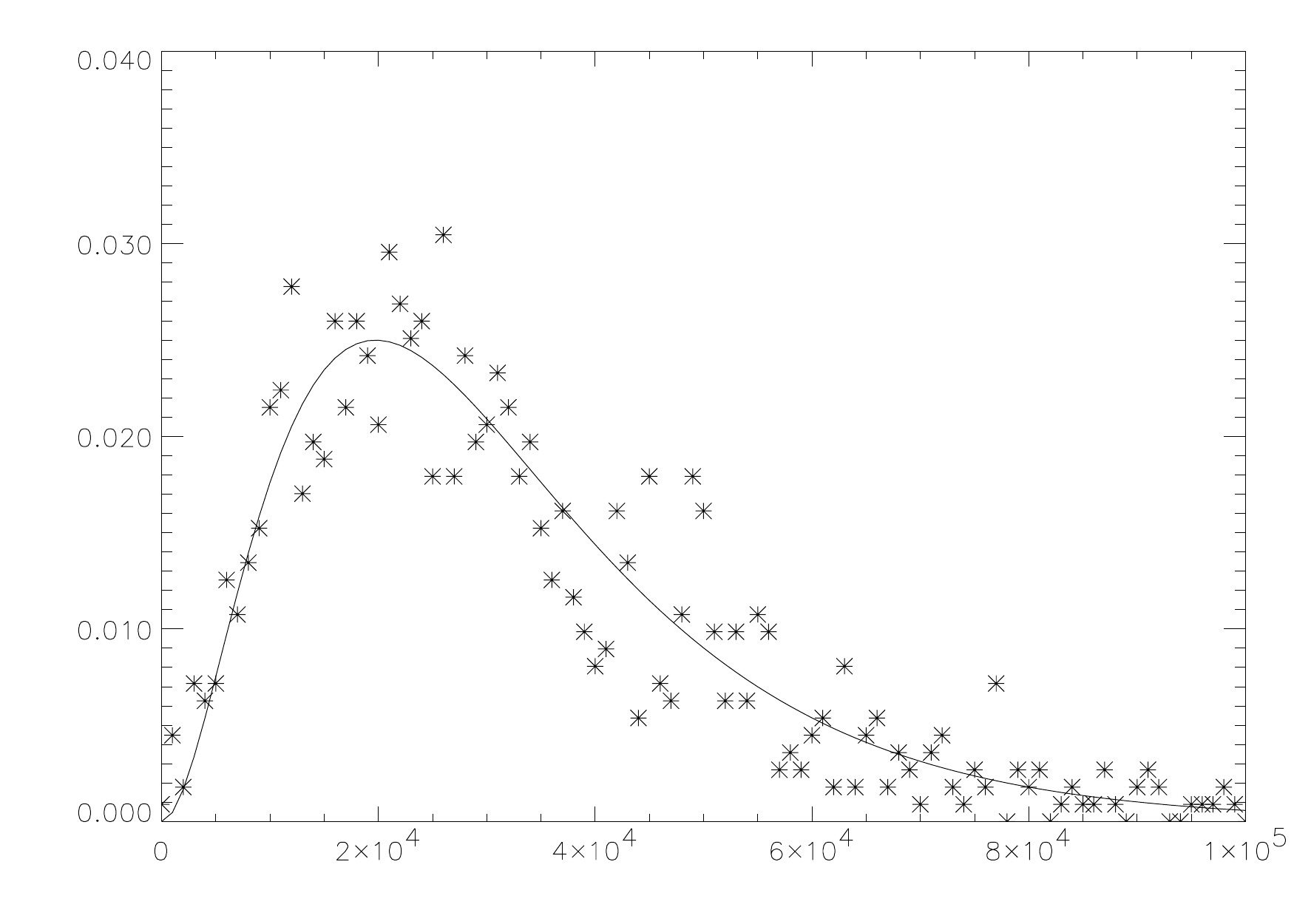}\label{fig_grass_hh_gh}}
\subfigure[HV $\mathcal{G}_{P}^{H}$]{\includegraphics[width=.32\linewidth]{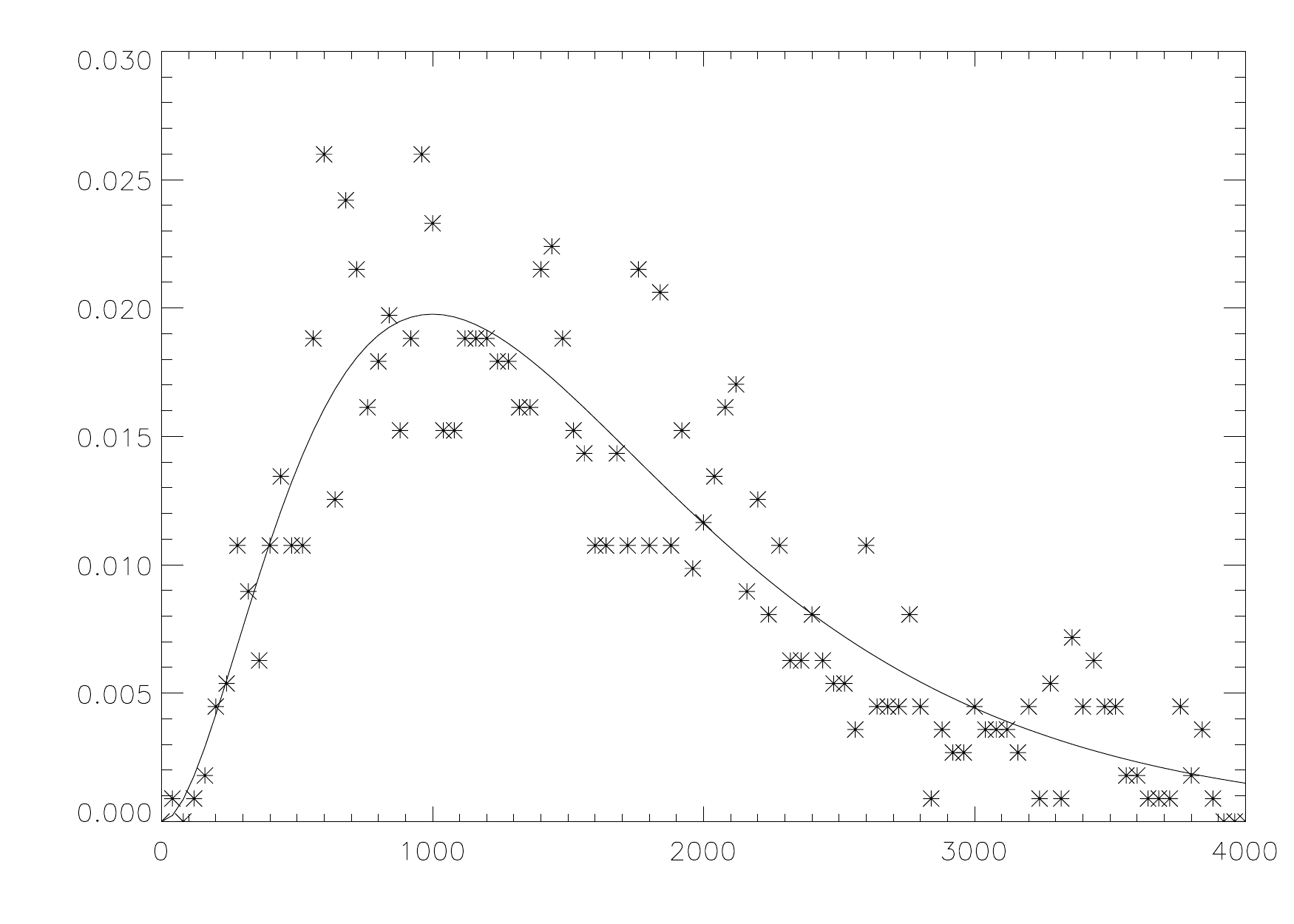}\label{fig_grass_hv_gh}}
\subfigure[VV $\mathcal{G}_{P}^{H}$]{\includegraphics[width=.32\linewidth]{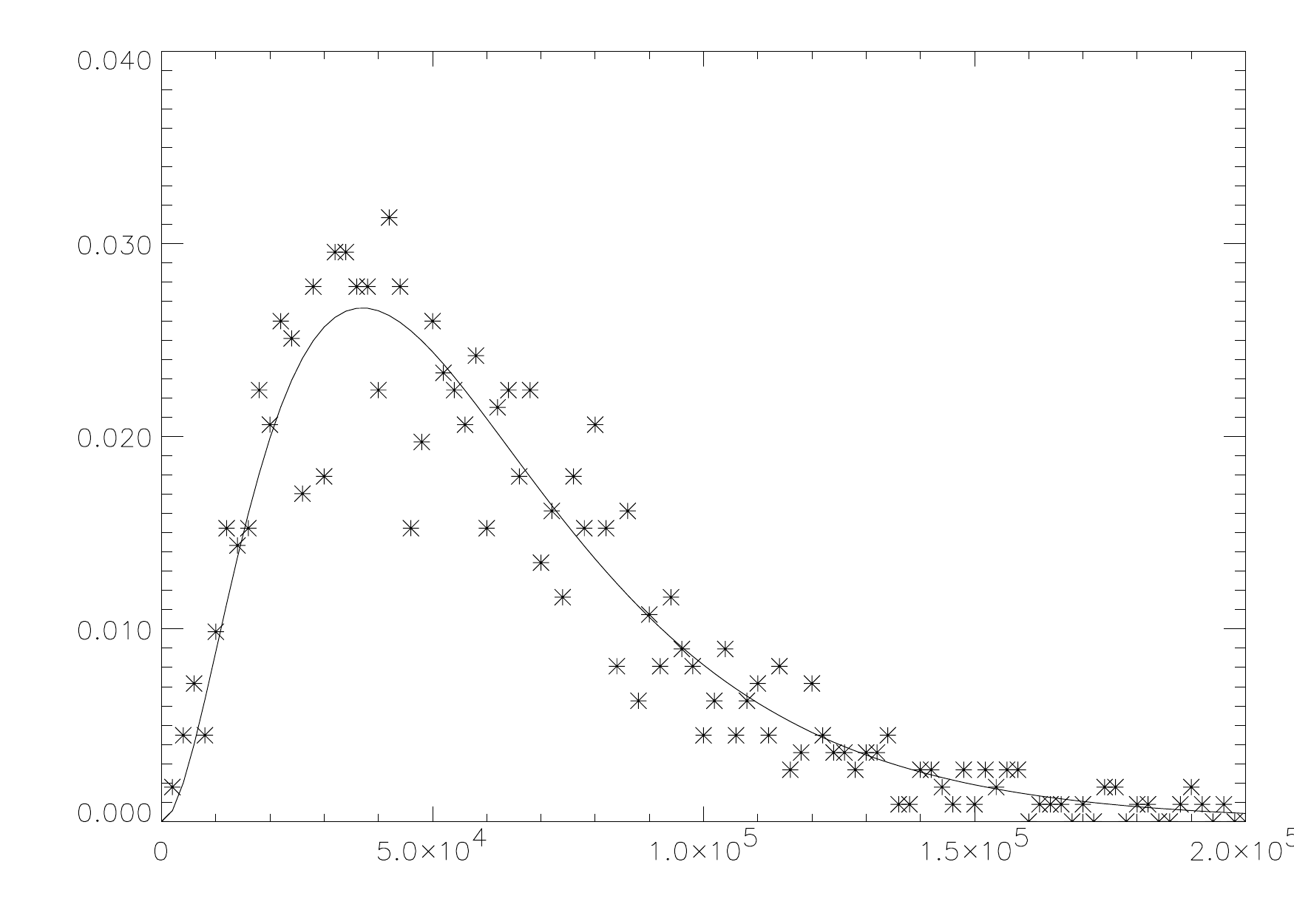}\label{fig_grass_v_gh}}
\caption{Histograms and estimated densities for pasture data}\label{fig_ajustes_grass}
\end{figure}

Figures~\ref{fig_ajustes_city},~\ref{fig_ajustes_forest} and~\ref{fig_ajustes_grass} show the histograms and estimated densities for different components and zones.
It is noticeable that the fit is excellent in urban samples and very good in the other cases.

\begin{figure}[hbt]
\centering
\subfigure[]{\includegraphics[width=.3\linewidth]{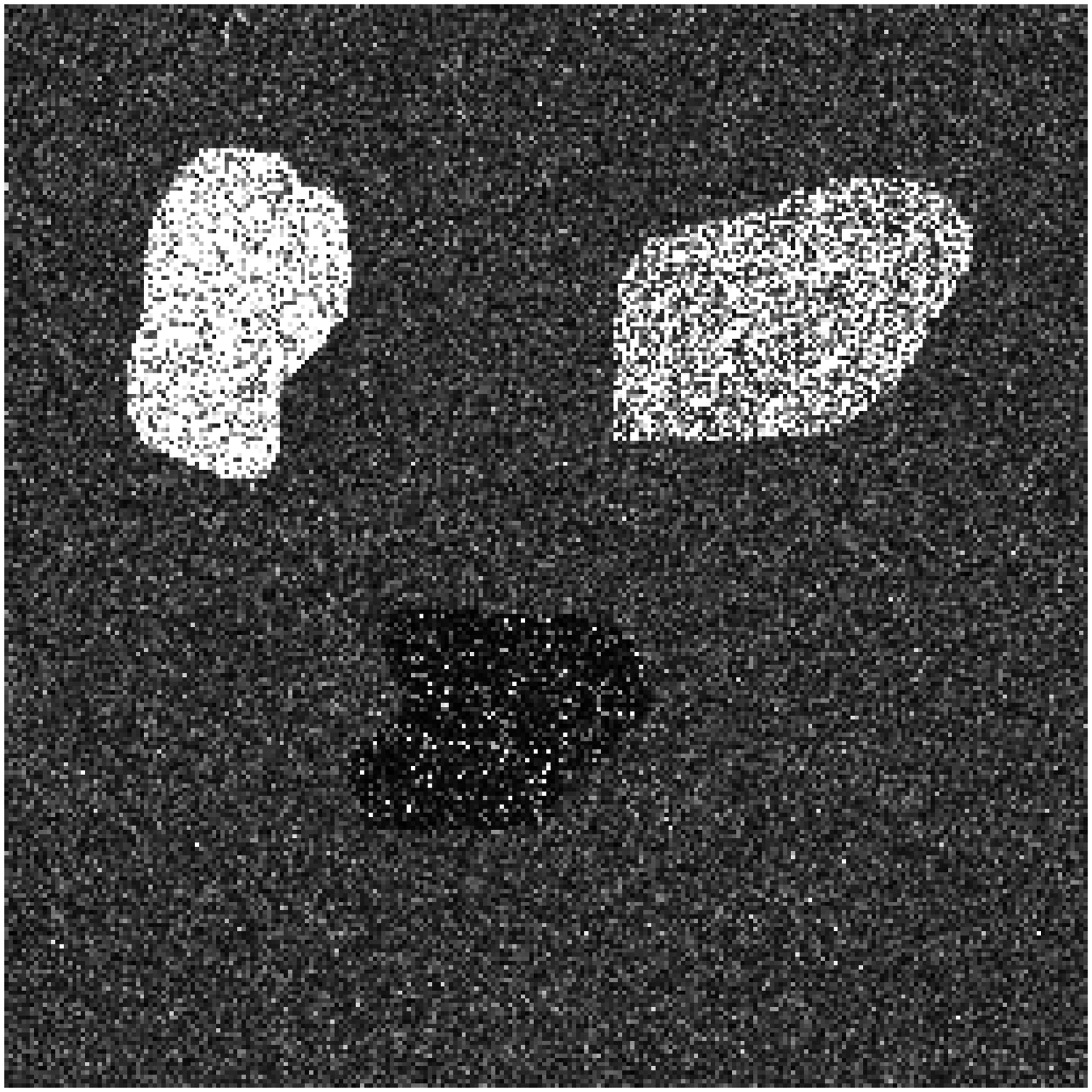}\label{fig_imag_hh}}
\subfigure[]{\includegraphics[width=.3\linewidth]{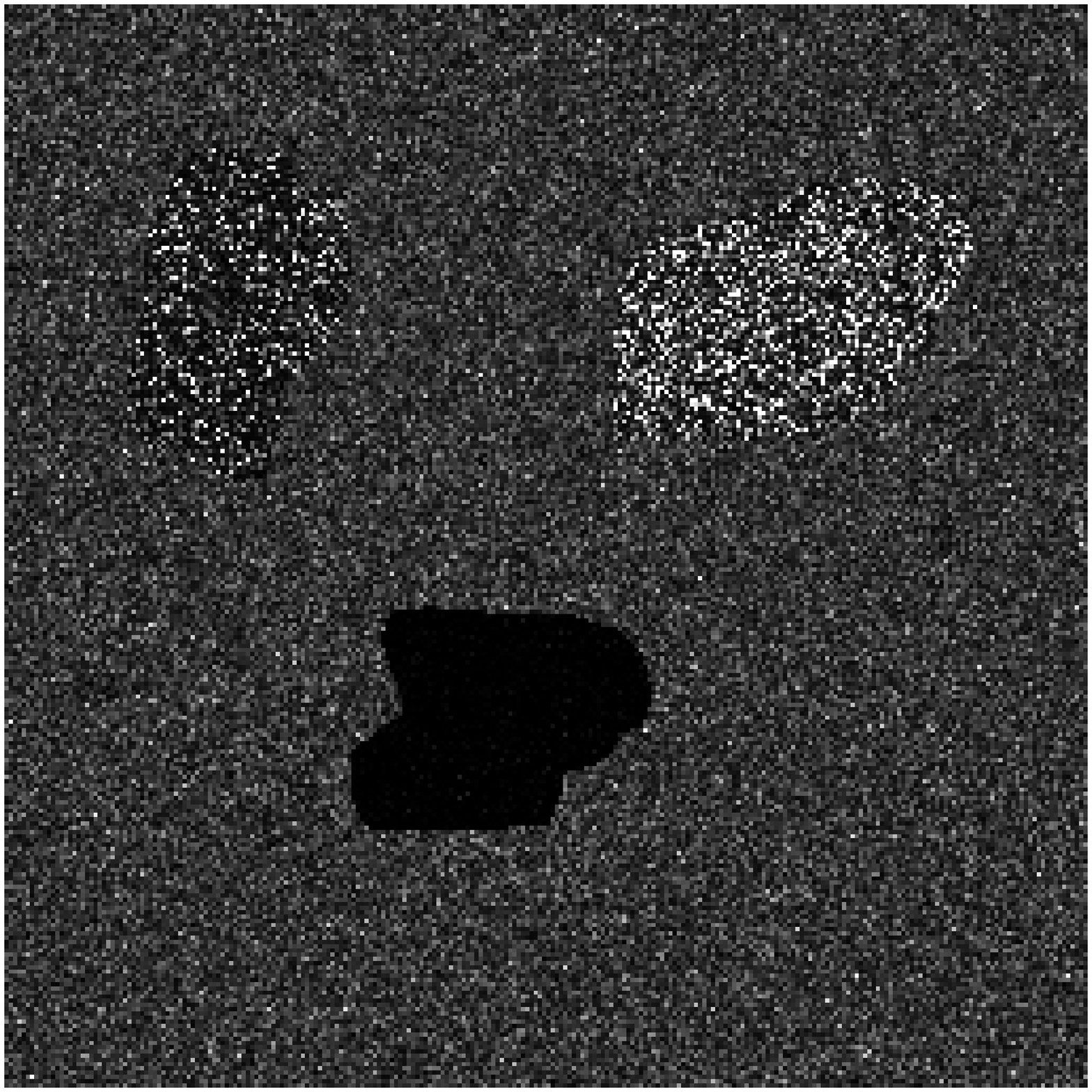}\label{fig_imag_hv}}
\subfigure[]{\includegraphics[width=.3\linewidth]{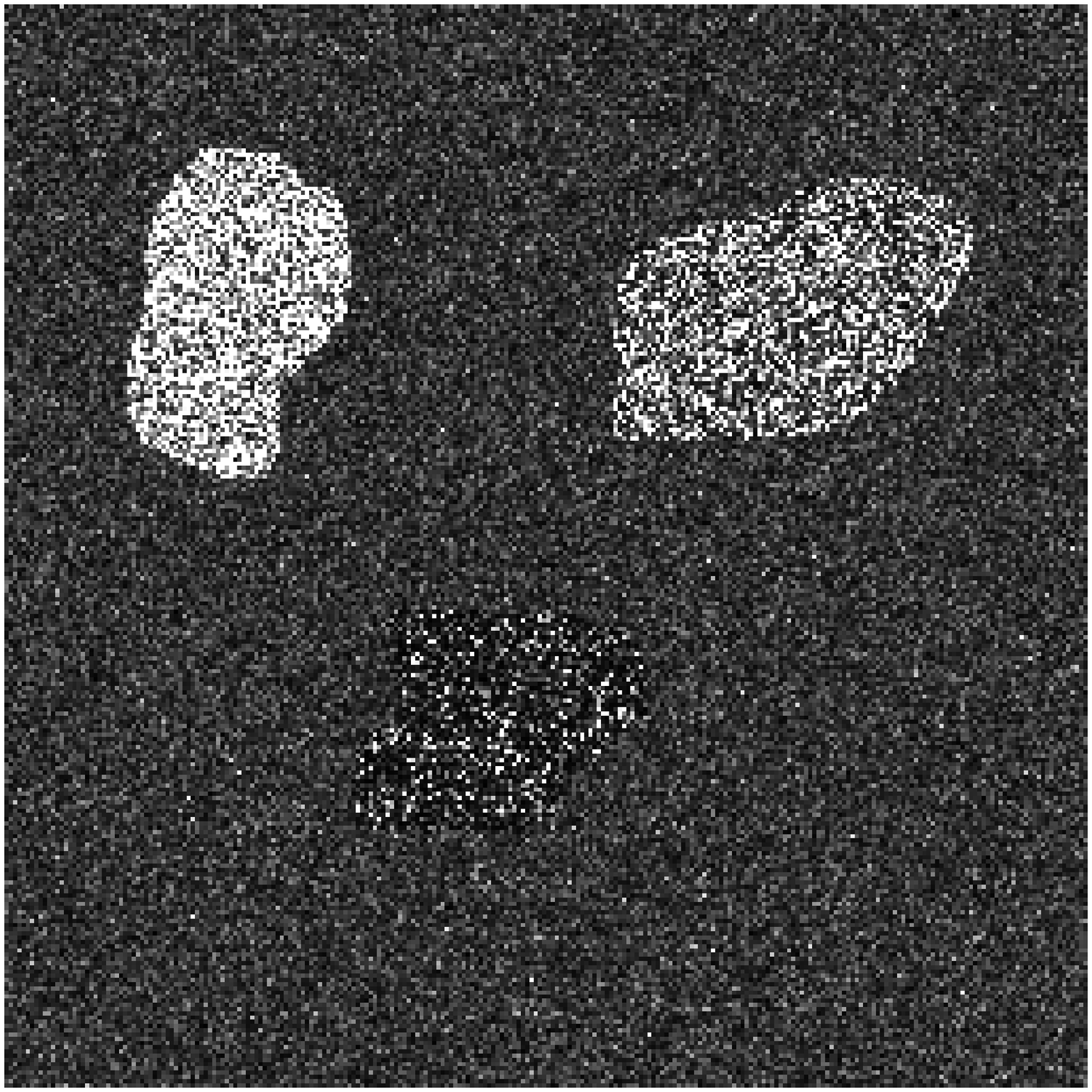}\label{fig_imag_vv}}
\subfigure[]{\includegraphics[width=.3\linewidth]{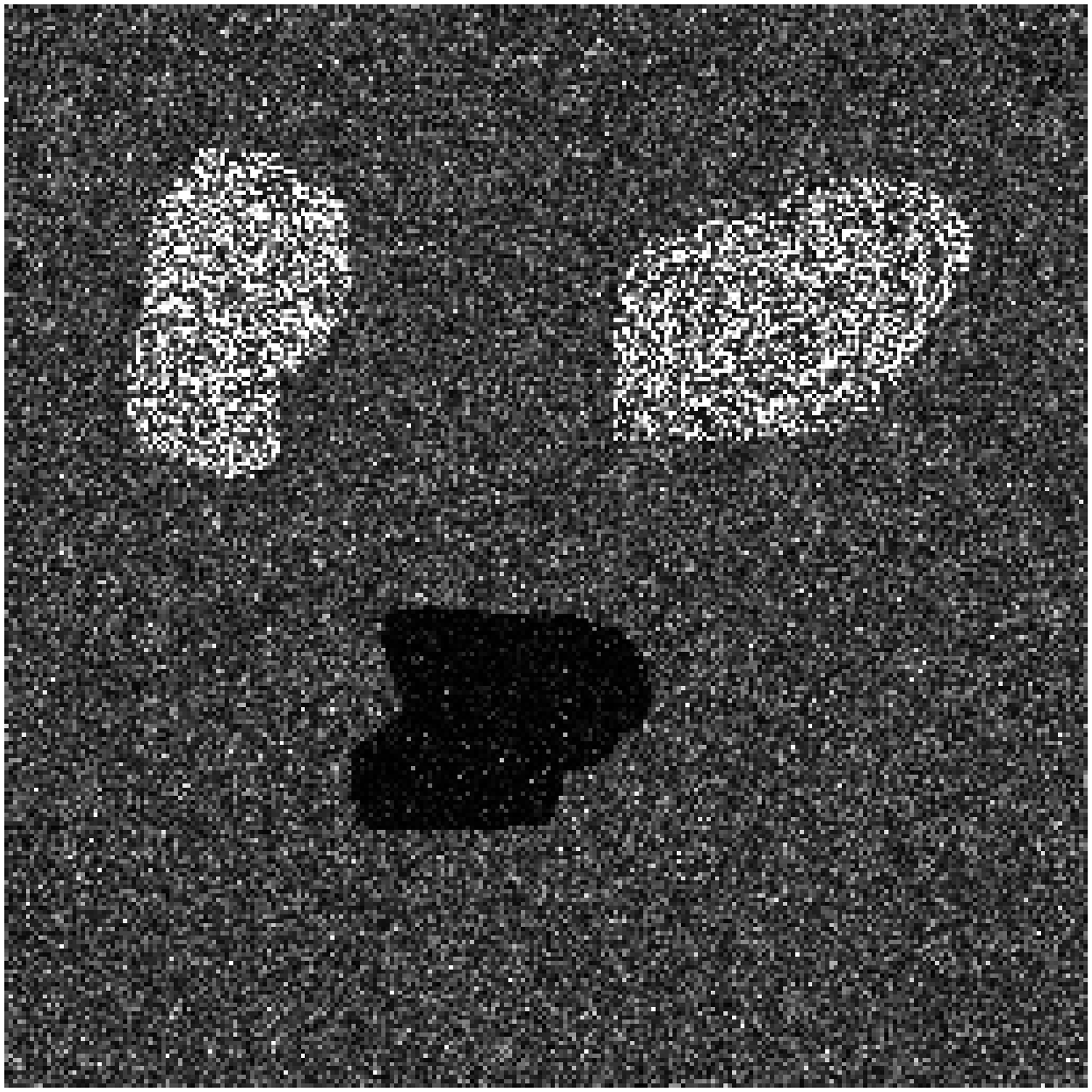}\label{fig_imag_hhhv}}
\subfigure[]{\includegraphics[width=.3\linewidth]{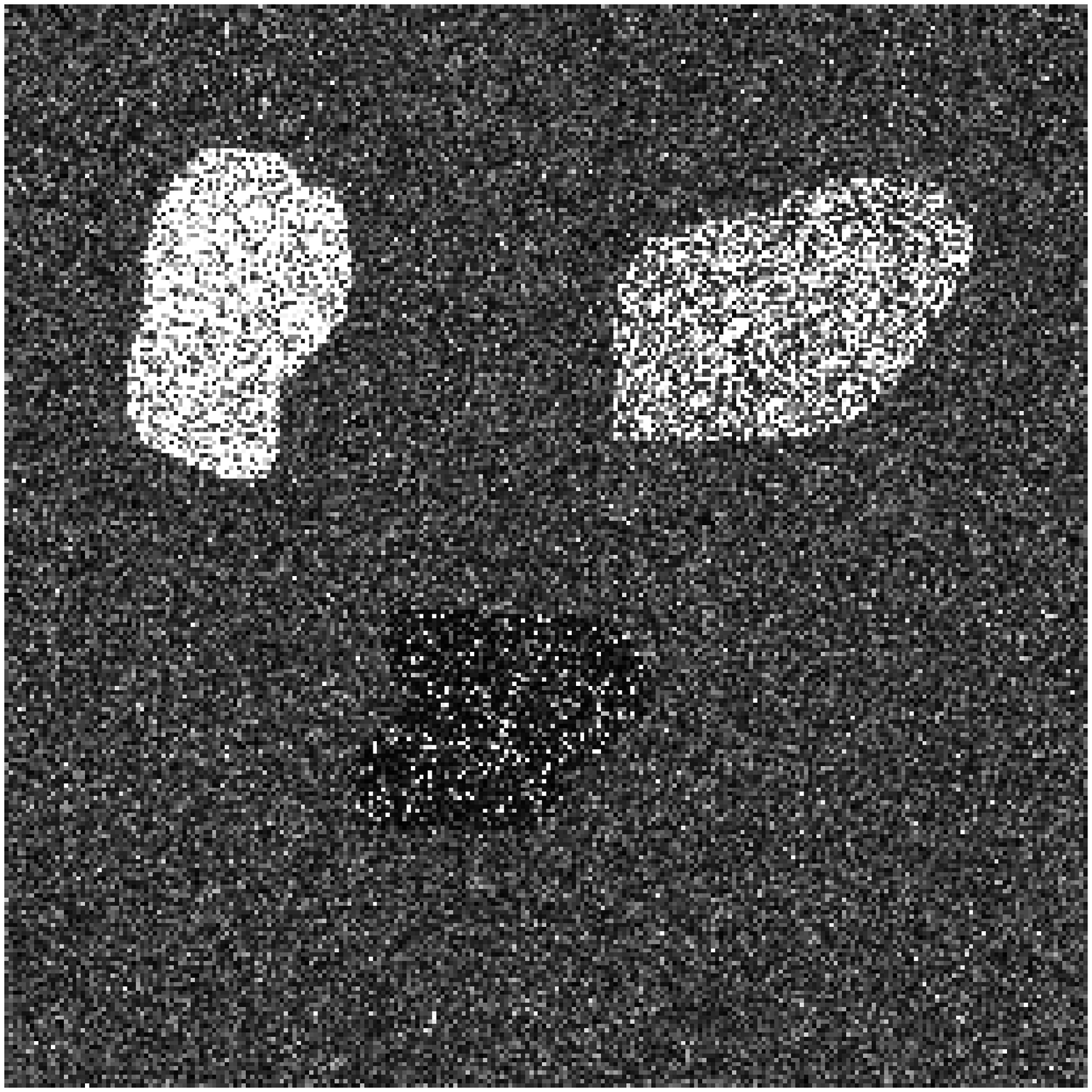}\label{fig_imag_hhvv}}
\subfigure[]{\includegraphics[width=.3\linewidth]{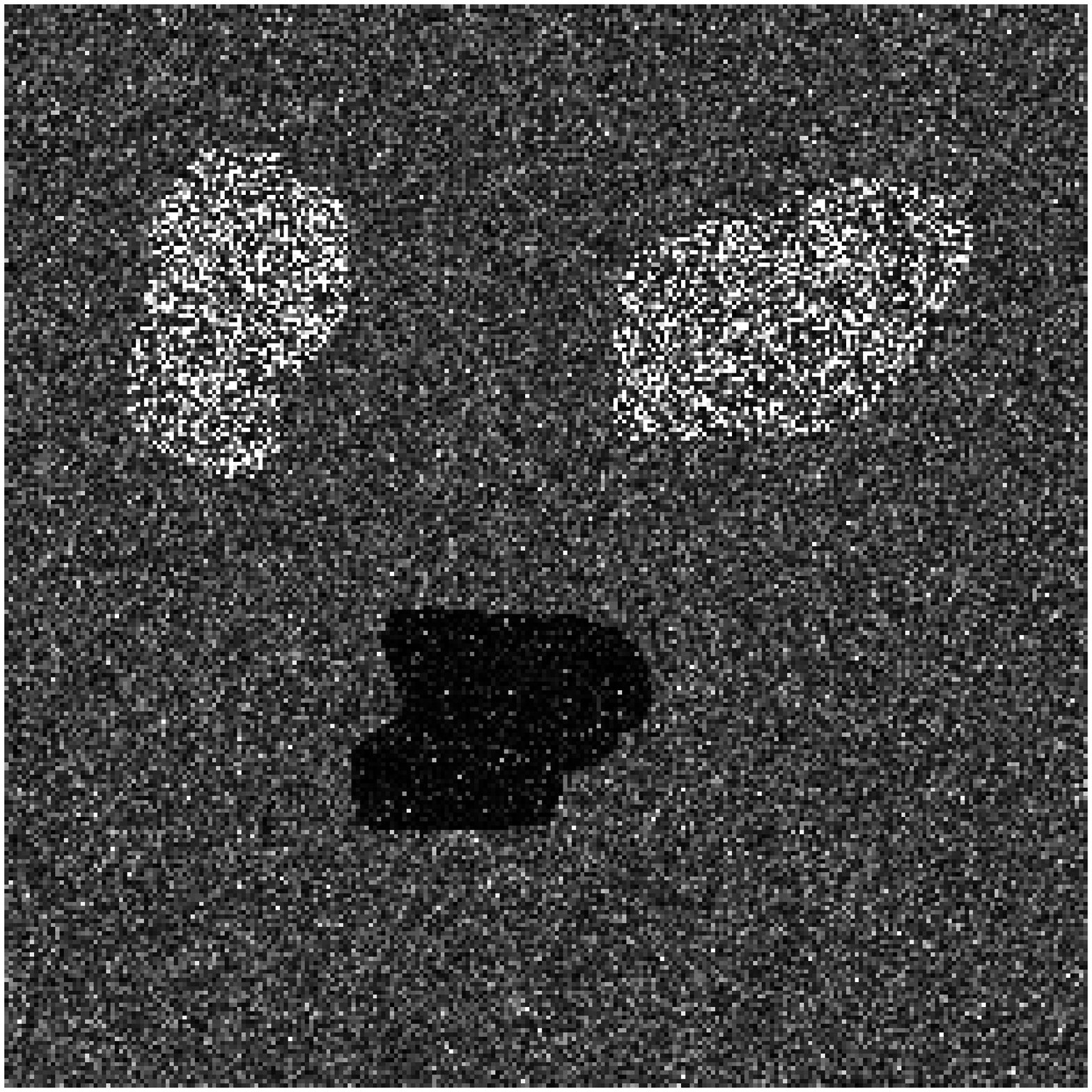}\label{fig_imag_hvvv}}
\caption{Synthetic images generated with different covariance matrices: a)~$| HH|^2$, b)~$| HV|^2$, c)~$| VV|^2$, d)~$| HH HV^*|$, e)~$| HH VV^*|$, f)~$| HV VV^*|$ }\label{fig_sinteticas_pol}
\end{figure}

Figure~\ref{fig_sinteticas_pol} shows synthetic images  with three regions generated with the estimated covariance matrices shown above.
The background in these figures has $\sigma^2_{HH}=\sigma^2_{HV}=\sigma^2_{VV}$ and no correlation among components.

The estimated values presented in equations~(\ref{matriz_sigmay_ciudad}), (\ref{matriz_sigmay_bosque}) and~(\ref{matriz_sigmay_pasto}), and in Table~\ref{tabla__omega} will be also used when assessing the error of the proposed technique.

\section{Boundary Detection}\label{boundary_detect}

In this section we describe an algorithm developed for boundary detection using B-Spline deformable contours.
Gambini et al.~\cite{Gambiniijrs} proposed this method univariate real SAR imagery, and here we adapt it to polarimetric data.

B-Splines are a convenient representation of spline functions  where the curve is specified by a few parameters, the control points; this reduces the computational effort to compute it.
The order of the polynomial segments is chosen arbitrarily, and it relates to the desired smoothness.
The B-Spline approach allows the local control of the curve by controlling the control points individually.
The curve lies within the convex hull induced by the control points.
For details of B-Spline representation of contours see the works~\cite{Blake,GambMejJacFreryStatistic,rogers90}.

A more sophisticated approach could be based, for example, on the minimum description length (MDL) principle~\cite{Figueiredo2000}, but the one presented here provided excellent results in real SAR applications.

\subsection{Initial Regions}

The procedure begins with a rough segmentation, either manual or automatic (with low computational cost), to be refined.

Let $E$ be a scene made up by the background $B$ and a region $ R$ with its  boundary $\partial R$.
We want to find a curve $C_B$ that fits $\partial R$ in the image.
We define first an initial search area, which is specified by polygons whose vertexes are the control points that generate a B-Spline curve. 
We developed an automatic algorithm for finding initial regions, but it also can be specified by the user.


If automatic initialization is chosen, the following process is performed.
The polarimetric image is of the form $f\colon S\rightarrow \mathbb{R}^6$, where $S=[0,\dots, ms_{b\ell}-1]\times [0,\dots, ns_{b\ell}-1]$, $m, n, s_{b\ell} \in \mathbb N$, i.e., it is composed of $m\times n$ blocks of side $s_{b\ell}$.
The following data entries have to be specified
\begin{enumerate}
\item A selection criterion $T_R \subset \mathbb{R}^+$ which depends on the homogeneity of the zone of interest. For example, in order to find urban areas we choose $T_R=[0.1, 1.5)$; this can be done using natural language and a table that converts text into parameter intervals.

\item A threshold $T_S$ that corresponds to the minimum number of blocks required for considering a candidate zone as an initial region. This specification can also be done in an intuitive and natural manner in terms of metric units provided the pixel resolution.
\end{enumerate}

The parameter $\omega$ is estimated for each block $S_{ij},\; i=0,\dots, m, \; j=0,\dots, n$, forming an array of size $m\times n$ of roughness estimates $\hat{\omega}(i,j)$.
Notice that each estimate is based on $s_{b\ell}^2$ samples.
We opted for $s_{b\ell}=11$, after experimenting different values in images with varying complexity; bigger windows provide more precise estimates when acting on areas evenly occupied, but will be more prone to mixing samples from different targets.
If $\hat{\omega}(i,j) \in T_R$, the block $S_{ij}$  is marked as candidate zone, else it is left unmarked.

If the number of connected blocks of a candidate zone is below $T_S$, then the zone is considered as noise and it is discarded.
The initial regions are formed by blocks, whose convex hull is then computed.


In this way, we define an initial search area  by means of the automatic determination of candidate connected components, which are specified by polygons whose vertexes are control points that generate a B-Spline curve.
Alternatively, the user is allowed to manually specify as many as desired regions of interest.

Once the initial search zones are determined, their centroids are calculated and the algorithm proceeds with the contour detection.

\subsection{Contour Detection}

If a point belongs to the object boundary, then a sample taken from the neighborhood of that point should exhibit a change in the statistical parameters.
We consider $N$ segments $s^{( i) }$, $i=1,\dots,N$ of the form $s^{( i) }=\overline{CP_{i}}$ for each candidate area, being $C$ the centroid of the initial region, the extreme $P_{i}$ a point outside of the region and $\theta =\angle( s^{(i)},s^{(i+1)}) $ the angle between two consecutive segments, for every $i$.
It is necessary for the centroid $C$ to be in the interior of the object whose contour is sought.
The points $P_{i}$, $i=1,\dots,N$ are arbitrarily chosen with the condition that they are outside the object of interest.

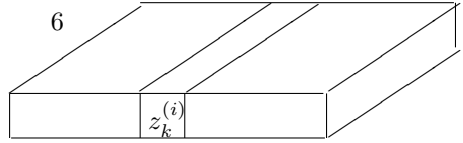
\begin{figure}[hbt]
	\begin{center}
		\setlength{\unitlength}{3158sp}%
\begingroup\makeatletter\ifx\SetFigFont\undefined%
\gdef\SetFigFont#1#2#3#4#5{%
  \reset@font\fontsize{#1}{#2pt}%
  \fontfamily{#3}\fontseries{#4}\fontshape{#5}%
  \selectfont}%
\fi\endgroup%
\begin{picture}(3542,1105)(2139,-1986)
\thinlines
\put(3176,-911){\line( 1, 0){2463}}
\put(2163,-1974){\line( 1, 0){2463}}
\put(2151,-1974){\line( 0, 1){350}}
\put(4639,-1974){\line( 0, 1){350}}
\put(2151,-1624){\line( 3, 2){1030.385}}
\put(4656,-1606){\line( 3, 2){1030.385}}
\put(4656,-1969){\line( 3, 2){1030.385}}
\put(2151,-1624){\line( 1, 0){2463}}
\put(5651,-1262){\line( 0, 1){350}}
\put(3176,-1974){\line( 0, 1){350}}
\put(3173,-1607){\line( 3, 2){1030.385}}
\put(3526,-1974){\line( 0, 1){350}}
\put(3543,-1613){\line( 3, 2){1030.385}}
\put(2476,-1124){\makebox(0,0)[lb]{\smash{{\SetFigFont{10}{12.0}{\rmdefault}{\mddefault}{\updefault}$6$}}}}
\put(3240,-1900){\makebox(0,0)[lb]{\smash{{\SetFigFont{10}{12.0}{\rmdefault}{\mddefault}{\updefault}$z_k^{(i)}$}}}}
\end{picture}%
	\end{center}
	\caption{Data structure for the segment $s^{(i)}$.}
	\label{caja}
\end{figure}

The segment $s^{( i) }$ is an array of $m \times 6$ elements coming from a discretization of the straight line on the array of the polarimetric image and is given by:
\[
s^{( i) }=\bigl( z_{1}^{( i) },\ldots ,z_{m}^{(
i) }\bigr),\quad 1\leq i\leq N,
\]
where $z_{k}^{(i)}, k=1,\dots, m$ is an array of $6$ elements as Figure~\ref{caja} shows.

In order to find the transition point on each segment $s^{(i)}$, the parameter $\omega$ is estimated as explained in section~\ref{estima_parametros} using a rectangle around the segment and a sliding window.
Then a set of estimates $\hat{\Omega}^{(i)} = (\widehat\omega_1, \dots, \widehat\omega_m)$ is obtained and the biggest variation of $\hat{\Omega}^{(i)}$ within the array is found, following Blake and Isard~\cite{Blake}, convolving it with an appropriate mask.
The coordinate at which $\widehat\omega$ exhibits the most intense variation is then considered a border point.
Once the set of border points $A= \{b_{1},\dots,b_{N}\}$ is found, the method builds the interpolating B-Spline curve.
Algorithm~\ref{boundary_detect_algor} shows a summary of the process to find the border points.

\begin{algorithm}[hbt]
\caption{Boundary Detection}\label{boundary_detect_algor}
\begin{algorithmic}[1]
\STATE Find or specify initial candidates and use the vertexes of the initial regions as control points of the starting boundary.
\STATE Determine a series of radial segments on the image.
\FOR{each segment}
	\STATE Generate sets of estimates of the parameter $\omega$ using the data within a rectangular window around the segment.
	\STATE Detect the border point by convolving each set with a border detection operator.
\ENDFOR
\STATE Return the B-Spline that interpolates the points found.
\end{algorithmic}
\end{algorithm}


\subsection{Error Evaluation}\label{error}

In this section we present the methodology employed for assessing the precision of the edge detection technique previously discussed.

In Polarimetric SAR images it is very difficult to evaluate the committed segmentation error due to the difficulty in establishing the true or correct segmentation.
Udupa et al.~\cite{Udupa2006} propose a framework for evaluating image segmentation algorithms and a systematic way to compare two different algorithms.
In order to obtain accuracy in error evaluation, the authors define the surrogate of truth in two possible ways, manual delineation and mathematical phantoms.
The first consists in tracing object boundaries manually, while the second consists in creating a set of as realistic as possible simulated images.
In this paper we use the second option, in order to measure de local error.
We successfully used this methodology for error evaluation in univariate amplitude SAR data \cite{GambMejJacFreryStatistic}.

\begin{figure}[hbt]
\begin{center}
\includegraphics[width=\linewidth,clip]{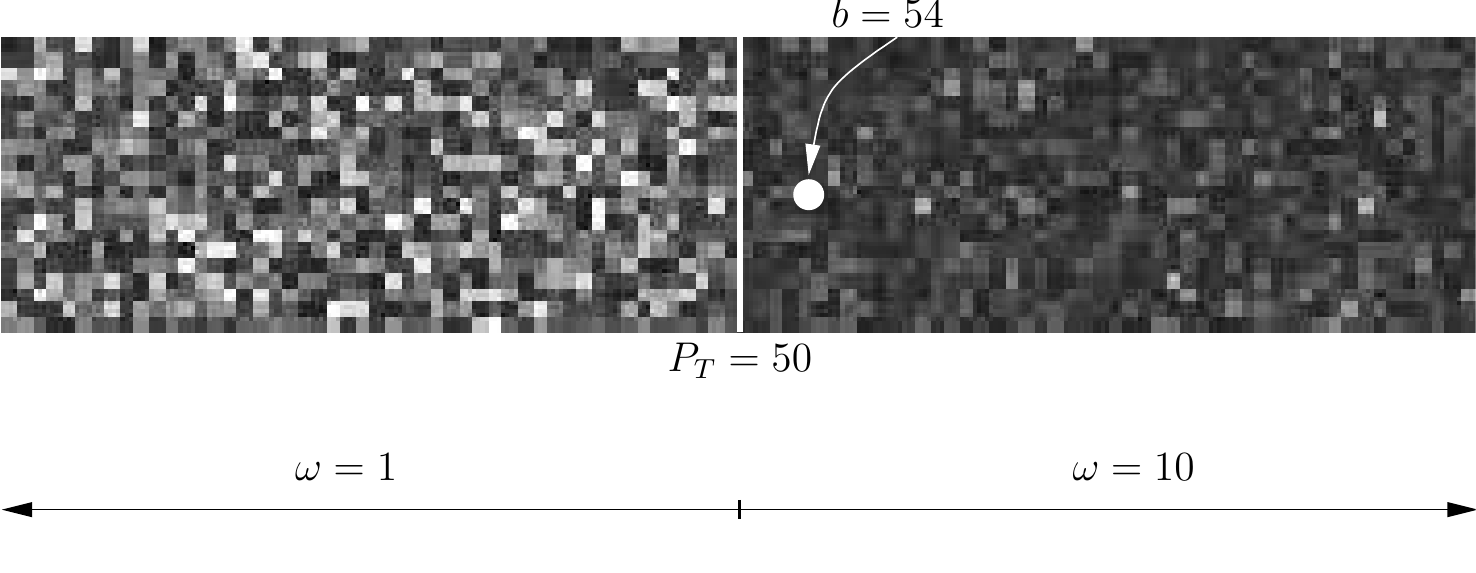}
\end{center}
\caption{Sample of the phantom image with true and estimated boundary points}\label{fig:tripa}
\end{figure}

A phantom image was used to compute the error in estimating by $b$ the true boundary point $P_T$.
This image is a $20\times 100$ pixels data set divided in halves, and each half is filled with samples from the $\mathcal{G}_H$ distribution with different parameters using the algorithm presented in \ref{app:simulation}.
Figure~\ref{fig:tripa} shows the one look situation, with $\omega=10$ to the right and $\omega=1$ to the left, both of them have the covariance matrix $\widehat\Sigma_{\text{u}}$, estimated using real data.
The figure also shows the correct border (the white vertical line at $50$) and an estimated transition point (the white dot denoted $b$ at $54$).
The error in this situation would be of four pixels.

%

Two hundred replications were made for each situation, the transition points were estimated and the error was estimated.
The distance of these points to the true boundary was evaluated and then the array $E$ was defined in each situation, as
\begin{equation}
E(r)= \left|50-b(r)\right|,\; 1\leq r\leq 200, \label{arreglo_A}
\end{equation}
where $b(r)$ is the transition point found in the $r$-th replication.

We use relative frequencies in order to estimate the probability of having an error smaller than a certain number of pixels.
Denote by $H(k)$ the number of replications for which the error is smaller than $k$ pixels, then an estimate of this probability is given by $f(k)={H(k)}/{200}$ for $k\geq 0$.
Algorithm~\ref{frecuencias} illustrates this process.
The bigger this probability, the better the algorithm.
Note that values of $k$ close to $0$ are the ones that must be taken into account to evaluate the technique under study; gross errors are not interesting.

\begin{algorithm}[hbt]
\caption{Boundary position error estimation}\label{frecuencias}
\begin{algorithmic}[1]
\FOR{each situation}
	\FOR{each $1\leq r\leq 200$}
		\STATE  Simulate a polarimetric SAR image of size $20\times100$, with two regions.
		\STATE Find the transition point on the major axis of the sample $b(r)$.
		\STATE Find the distance between the found and the true transition points.
		\STATE Update $E$, as defined in eq.~\eqref{arreglo_A}.
	\ENDFOR
	\STATE Compute $H$.
	\STATE Return the relative frequencies $f$.
\ENDFOR
\end{algorithmic}
\end{algorithm}

Twelve situations were considered in the study, each one referring to a pair of areas of different type modeled as
\begin{enumerate}
 \item $\widehat\Sigma_{\text{u}}$ and $\omega\in\{1,5\}$ besides $\widehat\Sigma_{\text{f}}$ and $\omega\in\{10,15\}$;
 \item $\widehat\Sigma_{\text{u}}$ and $\omega\in\{1,5\}$ besides $\widehat\Sigma_{\text{p}}$ and $\omega\in\{20,25\}$;
 \item $\widehat\Sigma_{\text{f}}$ and $\omega\in\{10,15\}$ besides $\widehat\Sigma_{\text{p}}$ and $\omega\in\{20,25\}$.
\end{enumerate}
The covariance matrices $\widehat\Sigma_{\text{u}}$, $\widehat\Sigma_{\text{f}}$ and $\widehat\Sigma_{\text{p}}$ are the ones estimated using real data an presented in equations~(\ref{matriz_sigmay_ciudad}), (\ref{matriz_sigmay_bosque}) and~(\ref{matriz_sigmay_pasto}), respectively.

\begin{figure}[hbt]
	\begin{center}
		\includegraphics[angle=-90,width=\linewidth]{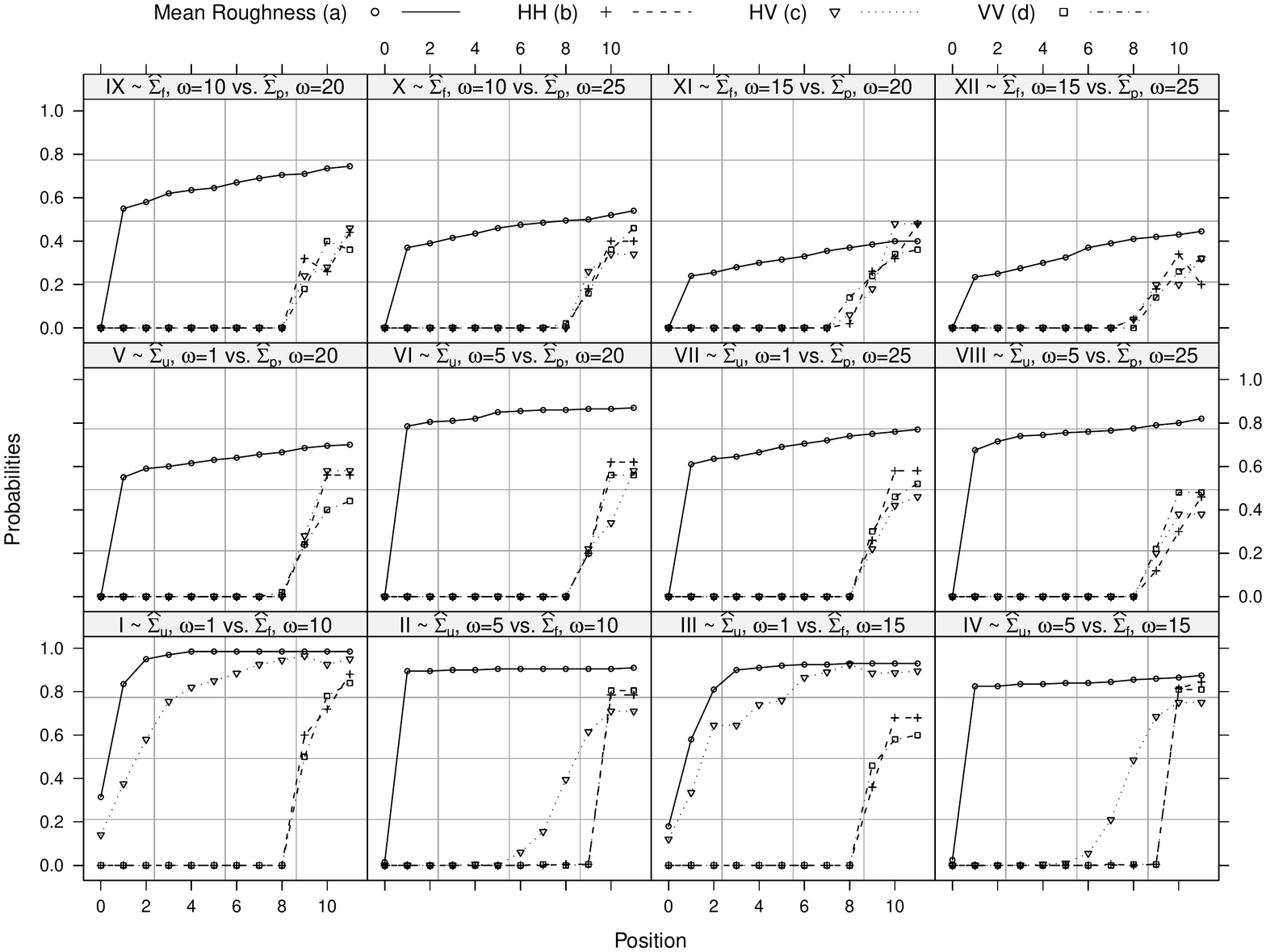}
	\end{center}
	\caption{Probability of finding the transition point with an error lower than the abscissa
	}	\label{graf_prob_gA0}
\end{figure}

Figure~\ref{graf_prob_gA0} shows the probability of finding the border point with an error lower than the number of pixels indicated on the horizontal axis for each of the twelve situations considered.
Four curves are shown for each situation: three discontinuous, related to single-channel data, and a continuous, which presents the results of using the mean of the estimated roughness parameters $\widehat\omega$ on the three channels.

The first conclusion is that using information from the three channels greatly enhances the discriminatory capability of the algorithm, in complete accordance with other results in the literature.
With the sole exception of situation XI, where every technique fails to detect the edge and we observe just random fluctuations, continuous lines are above the other ones.
Regarding the information content of each individual channel, HV polarization outperforms the other two: notice that the curve labeled~(c) is in most situations above curves~(b) and~(d).
In the following, we will only analyze the results obtained with the three channels, i.e., continuous lines labeled~(a).

It is noticeable that urban patches can be easily discriminated from forests and from pasture, since all situations but X, XI and~XII, where there no urban data are used, rapidly rise to values close to $1$.
The discrimination between pasture and forest only is a hard task, c.f. situations X, XI and~XII, being the first, i.e., $\widehat\Sigma_{\text{f}}, \omega=10$ against $\widehat\Sigma_{\text{p}}, \omega=25$ the only feasible among them.

\subsection{Application to real data}\label{sec_results}

\begin{figure}[hbt]
	\begin{center}
\includegraphics[width=\linewidth]{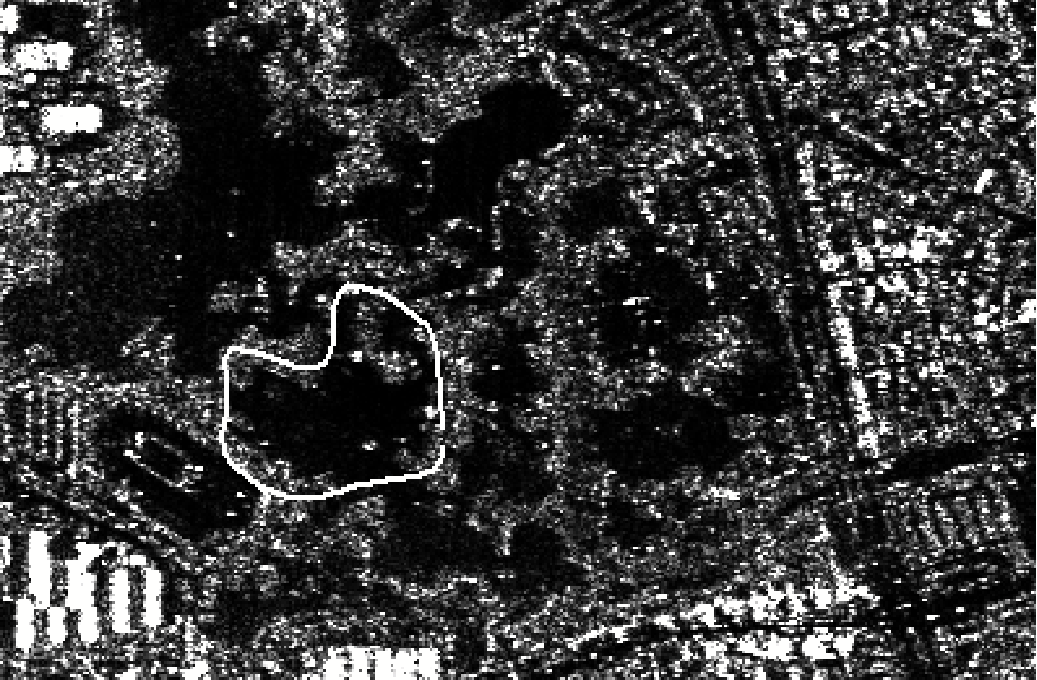}
	\end{center}
	\caption{Real polarimetric three looks E-SAR image and boundary detection with automatic initialization}\label{fig_imag_real:b}
\end{figure}

%
%
Figure~\ref{fig_imag_real:b} shows the HH band of a $3$ looks real E-SAR image showing an urban area from the city of Munich, along with a region boundary detected using Algorithm~\ref{boundary_detect_algor}.
The control parameters specified a medium size homogeneous area and, as can be seen, the technique deals well with both complex structures and noisy data.

\begin{figure}[hbt]
	\begin{center}
		\includegraphics[width=.8\linewidth]{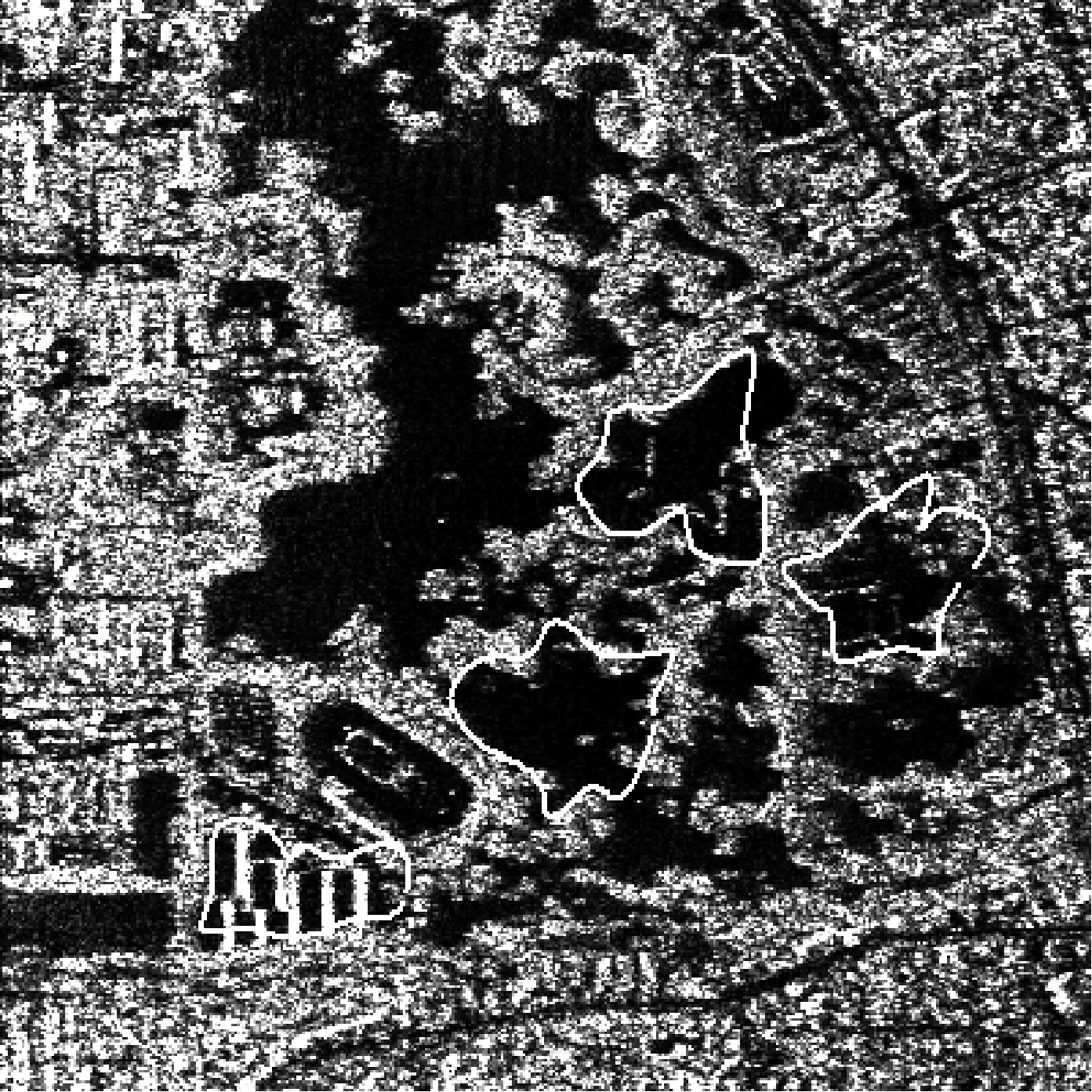}
	\end{center}
	\caption{Result of applying Algorithm~\ref{boundary_detect_algor} to a $3$-looks polarimetric image with four initial regions manually specified}
	\label{fig_imag_real_muchas_regiones}
\end{figure}

Figure~\ref{fig_imag_real_muchas_regiones} shows the result of applying Algorithm~\ref{boundary_detect_algor} to the same image, but specifying four regions of interest manually.

Smooth curves were sought in both cases.
This was specified setting the degree of the polynomials to four.

Our proposal employs a statistical model with interpretable parameters, leading to more information than just the detected edges and the usual geometrical features (area, shape etc.).
In the case of figure~\ref{fig_imag_real:b} the algorithm informs that $\widehat\omega=17.32$; the detected area conforms, then, to our requirement of being homogeneous.
The four regions found in figure~\ref{fig_imag_real_muchas_regiones} returned, from left to right, the following roughness estimates: $\widehat\omega=0,75$, $\widehat\omega=18,94$, $\widehat\omega=16,08$ and $\widehat\omega=12,88$.
Those areas are thus labeled as extremely heterogeneous, two homogeneous and one homogeneous tending to heterogeneous, respectively.

\section{Conclusions}\label{conclusiones}

Polarimetric SAR imagery segmentation is a very difficult task to solve.
In this work we described a new approach to region boundary detection in polarimetric SAR images using B-Spline deformable contours and local parameter estimation.
The boundaries of several regions with varying degrees of complexity were obtained using our proposal.

In the first step we either find or specify regions of interest that correspond to areas with different degrees of homogeneity, as a coarse first approximation.
For each region, its boundaries are considered as the initial solution for the border detector.
Then, the estimated parameter of roughness is calculated using two samples: one included in the region and the other out of the region and we find the transition point only for the data that are on a set surrounding a line segment.
All these processes diminish the computational cost and improve the performance of the method.

For each region, the result of the application of this algorithm is a boundary curve given by an expression in terms of B-Spline functions.
The results using both simulated and real SAR images are excellent and were obtained with an acceptable computational effort.

In addition, the error in finding edges was defined, and a Monte Carlo experiment was used to assess this error in a variety of situations that appear in the practice of polarimetric SAR imagery analysis.
The information content was quantified, in the sense that these results show that using the estimator of $\omega$ computed from the three intensity components consistently leads to better results than employing one polarization.

We did not consider the other nine parameters of the polarimetric model for the detection of transition points, given the very good performance obtained using the estimation of the $\omega $ alone.

Future work includes the use of more parameters for finer detail detection, improved and robust estimators (as, for instance, the ones presented in \cite{AllendeFreryetal:JSCS:05,%
BustosFreryLucini:Mestimators:2001,%
CribariFrerySilva:CSDA,%
FreryCribariSouza:JASP:04,%
asp98,%
SilvaCribariFrery:ImprovedLikelihood:Environmetrics,%
VasconcellosFrerySilva:CompStat}) and other types of deformable contour methods based on level sets and on stochastic distances \cite{HypothesisTestingSpeckledDataStochasticDistances}.

\section*{Acknowledgments}
We thank DLR for the E-SAR image provided, CNPq, FAPEAL and SeCyT for partial funding.

\appendix

\appsection{Intensity Univariate Distributions for SAR Data}\label{app:mono}
The multi-look return in intensity univariate SAR images can be modeled as the product of two independent random variables, one corresponding to the backscatter $X$ and other to the
speckle noise $Y$.
In this manner $Z=X \cdot Y $ models the return $Z$ in each pixel under the multiplicative model.
For univariate intensity data, the speckle noise $Y$ is modeled as a $\Gamma ( n,n) $ distributed random variable, where $n$ is the number of looks, while the backscatter $X$ is considered to obey a Generalized Inverse Gaussian law, denoted as $\mathcal{N}^{-1}( \alpha ,\lambda ,\gamma )$.

For particular values of the parameters of the $\mathcal{N}^{-1}$ distribution, the $\Gamma ( \alpha ,\lambda ) $, the $\Gamma ^{-1}( \alpha ,\gamma ) $, and the $IG( \gamma ,\lambda ) $ (Inverse Gaussian) distributions are obtained.
These, in turn, give rise to the $K$, the $\mathcal{G}^{0}$, and the $\mathcal{G}^{H}$ distributions for the return $Z$, respectively.

Given the mathematical tractability and descriptive power of the $\mathcal{G}^{0}$ (c.f. references \cite{MejailJacoboFreryBustos:IJRS,mejailfreryjacobobustos2001,QuartulliDatcu:04}) and the $\mathcal{G}^{H}$ distributions, they represent an attractive choice for SAR data modeling.
As in this work we will use the $\mathcal{G}^{H}$ we will describe it in more detail here.

The density of the Generalized Inverse Gaussian distribution is given by:
\begin{equation*}
f_{X}(x)=\frac{\left(  \lambda/\gamma\right)  ^{\alpha/2}}{2K_{\alpha}\left(
\sqrt{\lambda\gamma}\right)  }x^{\alpha-1} \exp\Bigl\{   -\frac{1}{2}\left(
\lambda x+\frac{\gamma}{x}\right) \Bigr\}  \mathbf{1}_{\mathbb{{{R}^{+}}}%
}\left(  x\right)  ,
\label{densGIGconGammaLambda}%
\end{equation*}
with parameters $\gamma$, $\lambda$ and $\alpha$ in the following parameter space:
\begin{equation*}
\left\{
\begin{array}
[c]{ccccc}%
\gamma>0 & \text{\textrm{and}} & \lambda\geq0 & \text{\textrm{if}} & \alpha<0\\
\gamma>0 & \text{\textrm{and}} & \lambda>0 & \text{\textrm{if}} & \alpha=0\\
\gamma\geq0 & \text{\textrm{and}} & \lambda>0 & \text{\textrm{if}} & \alpha>0,
\end{array}
\right.  \label{RangosParametrosAlphaGammaLambda}%
\end{equation*}
where
\begin{eqnarray*}
        \mathbf{1}_{A}( x)=\left\{ \begin{array}{ll}
                    1 & \mbox{if $x \in A$ }\\
                    0 & \mbox{if $x \notin A$},
                     \end{array} \right.
\end{eqnarray*}
and where $K_{\alpha}$ denotes the modified Bessel function of the third kind and order $\alpha$, given by
\[
K_{\alpha}(\sqrt{ab})=\left(  \frac{a}{b}\right)  ^{\alpha/2}\frac{1}{2}%
\int_{\mathbb{{{R}_{+}}}}x^{\alpha-1}\exp\left(  -\frac{1}{2}\left(
ax+bx^{-1}\right)  \right)  .
\]

The Inverse Gaussian distribution  $IG(\gamma,\lambda)$ is obtained when $\alpha=-1/2$, and  its density function is given by Eq.~(\ref{densGIconGammaLambda}):
\begin{equation}
f_{X}(x)=\sqrt{\frac{\gamma}{2\pi x^{3}}} \exp\Bigl\{-\frac{(
\sqrt{\lambda}x-\sqrt{\gamma})  ^{2}}{2x}\Bigr\}  \mathbf{1}%
_{\mathbb{{{R}^{+}}}}\left(  x\right)  ,
\label{densGIconGammaLambda}%
\end{equation}
with $\lambda,\gamma>0$.
The formula for the moments of this distribution is
\begin{equation*}
\mathbb{E}\left[  X^{r}\right]  =\left(  \frac{2}{\pi}\sqrt{\gamma\lambda
}\right)  ^{1/2}\exp(\sqrt{\gamma\lambda})\left(  \sqrt{\frac{\gamma}{\lambda
}}\right)  ^{r}K_{r-\frac{1}{2}}\left(  \sqrt{\gamma\lambda}\right)
,\label{momGIconGammaLambda}%
\end{equation*}
so, the moments of first and second order, and the variance are
$$
\mathbb{E}\left[X\right] =\sqrt{\frac{\gamma}{\lambda}},
\mathbb{E}\left[X^{2}\right] =\frac{\gamma}{\lambda}+\sqrt{\frac{\gamma}{\lambda^{3}}},\text{ and } 
\mathbb{E}\left[\left(X-\mathbb{E}\left[X\right]\right)^{2}\right] =\sqrt{\frac{\gamma}{\lambda^{3}}},
$$
respectively.

\begin{figure}[hbt]
\begin{center}
\includegraphics[width=12.0cm]{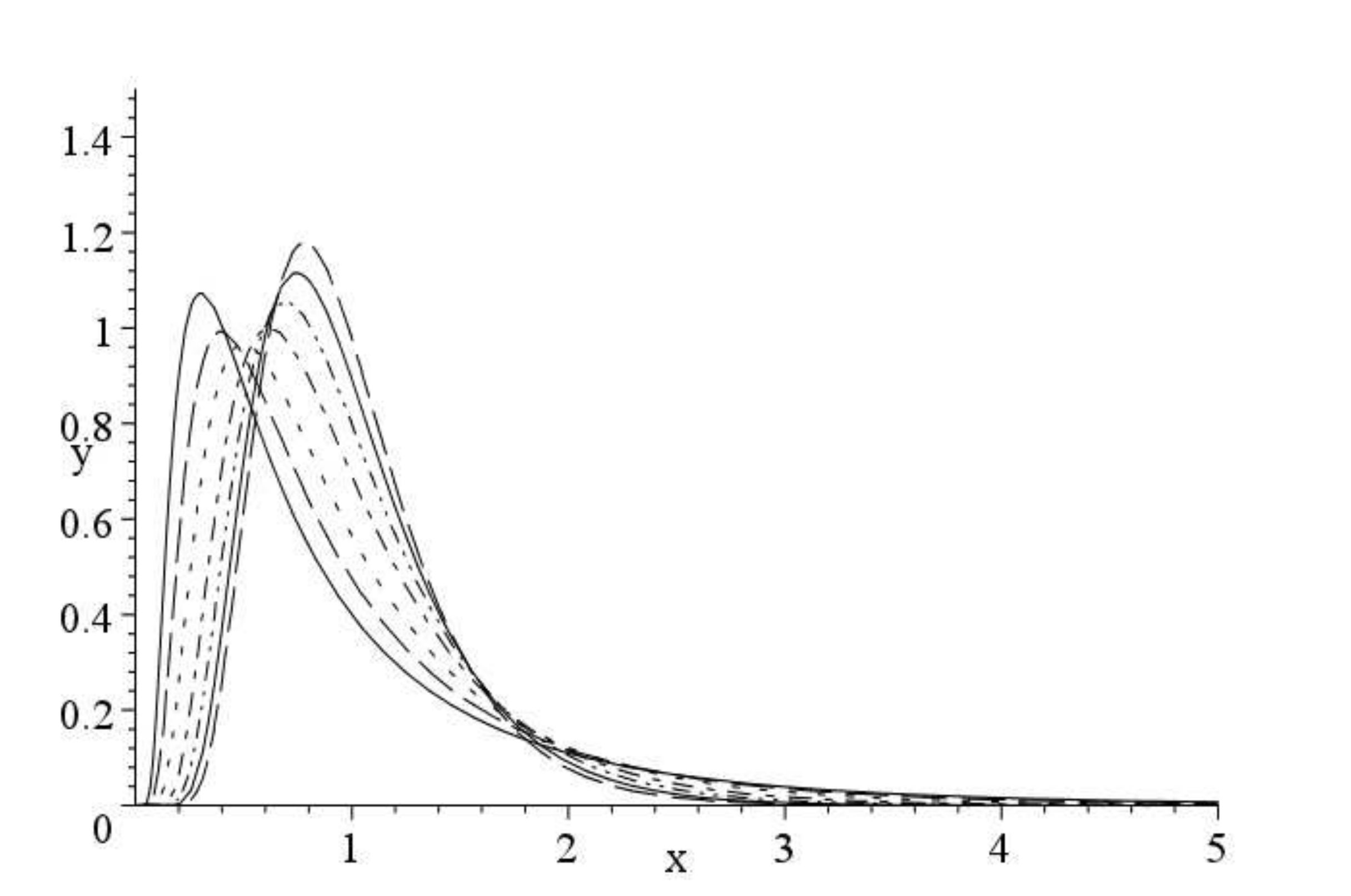}%
\caption{\small{Density of the $IG\left(  x,\omega,\eta\right)  $ distribution,
for $\eta=1$ and $\omega=1$ (solid), $\omega=\sqrt{2}$ (dashed), $\omega=2$ (dotted),
$\omega=3$ (dot-dash), $\omega=4$ (dot-dot-dash), $\omega=5$ (solid) and $\omega=6$ (dashed).}}%
\label{PlotDensGIconEtaFijoXX}%
\end{center}
\end{figure}

The parameters $\gamma$ and $\lambda$ can be used to define a new pair of parameters $\omega$ and $\eta$, using $\omega =\sqrt{\gamma\lambda}$ and $\eta =\sqrt{\gamma/\lambda}$, so equation~(\ref{densGIconGammaLambda}) can be rewritten as
\begin{equation*}
f_{X}(x)=\sqrt{\frac{\omega\eta}{2\pi x^{3}}}\exp\left(  -\frac{\omega}{2}%
\frac{\left(  x-\eta\right)  ^{2}}{x\eta}\right)  \mathbf{1}%
_{\mathbb{{{R}^{+}}}}\left(  x\right)  .\label{densGIconOmegaEta}%
\end{equation*}
Then, if $X\sim IG\left(  \omega,\eta\right)  $  it is possible to see that the corresponding moments are
\begin{equation*}
\mathbb{E}\left[  X^{r}\right]  =\sqrt{\frac{2\omega}{\pi}}e^{\omega}\eta
^{r}K_{r-\frac{1}{2}}\left(  \omega\right)  ,\label{momGIconOmegaEta_cap3}%
\end{equation*}
so the first and second order moments, and the variance are
$$
\mathbb{E}\left[X\right] =\eta,\label{mom1erOrdenGIconOmegaEta},
\mathbb{E}\left[X^{2}\right]  =\eta^{2}\frac{\omega+1}{\omega}, \text{ and }
\mathbb{E}\left[\left(X-\mathbb{E}\left[X\right]\right)^2\right]  =\eta^{2}\frac{1}{\omega},
$$
respectively.
With this re-parametrization, $\eta$ is the mean value and the variance grows as $\omega\to0$.
If $X\sim IG\left(  \omega,\eta\right)  $ then
$X/\eta\sim IG\left(  \omega,1\right)  $, and thus the density function of the random variable $X/\eta$ is given by Eq.~(\ref{densidad_de_X_sobre_eta})
\begin{equation}
f_{X\mid\eta}\left(  x\right)  =\sqrt{\frac{\omega}{2\pi x^{3}}}\exp\Bigl\{
-\frac{\omega}{2}\frac{(x-1)  ^{2}}{x}\Bigr\}  \mathbf{1}%
_{\mathbb{{{R}^{+}}}}\left(  x\right)  \label{densidad_de_X_sobre_eta}
\end{equation}
In Figure~\ref{PlotDensGIconEtaFijoXX} the curves corresponding to this density for $\eta=1$ and various values of $\omega$, are shown.
It is noticeable that the variance grows as the parameter $\omega$ approaches $0$.

\begin{figure}[hbt]
\begin{center}
\includegraphics[width=12.0cm]{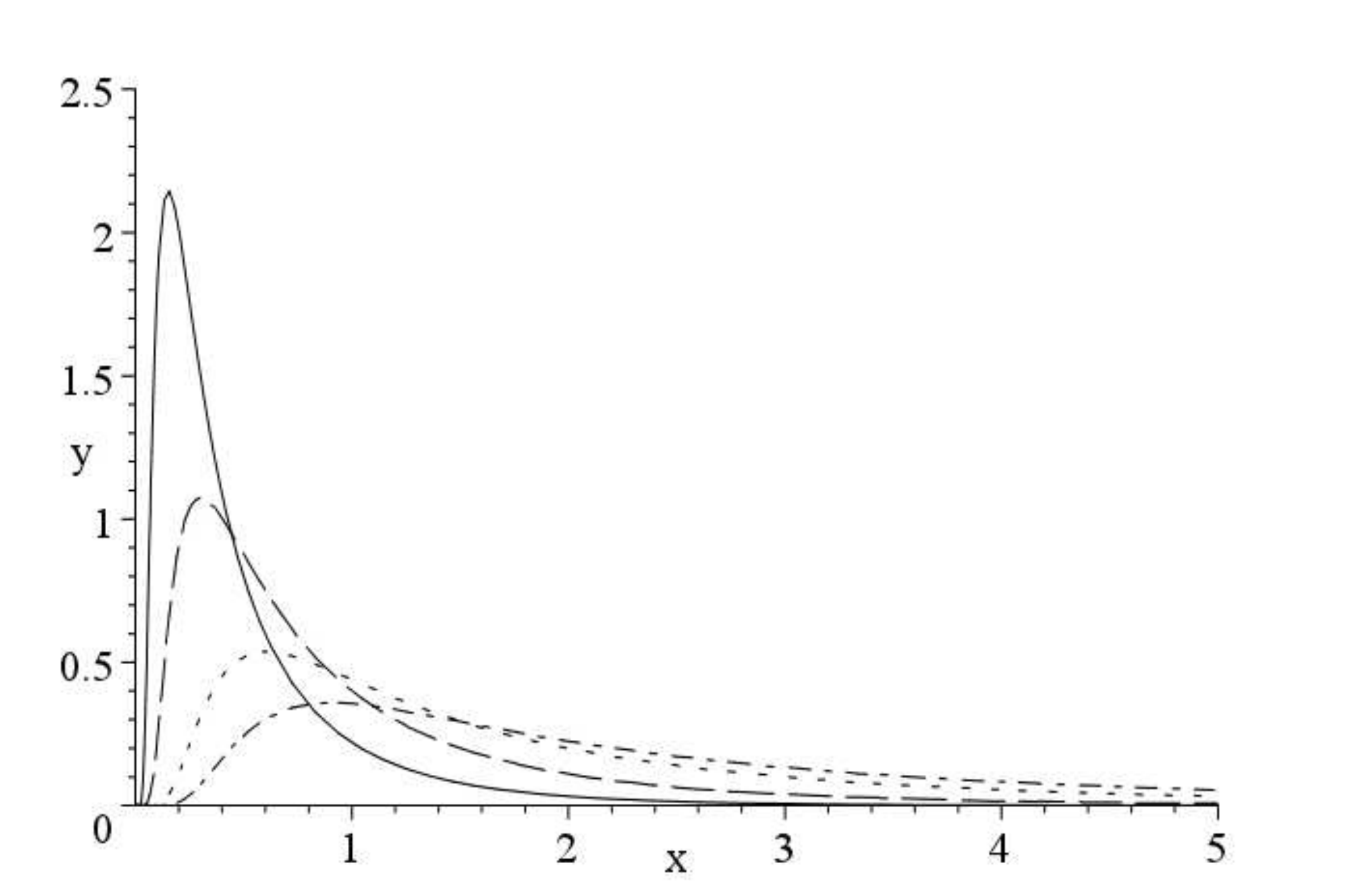}%
\caption{\small{Density of the $IG\left(  x,\omega,\eta\right)  $ distribution,
for $\omega=1$ and $\eta=0.5$ (solid), $\eta=1$ (dashed), $\eta=2$ (dotted) y $\eta=3$
(dot-dash).}}%
\label{PlotDensGIconOmegaFijo}%
\end{center}
\end{figure}

Figure~\ref{PlotDensGIconOmegaFijo} exhibits the curves for $\omega=1$ and various values of $\eta$.  Here, the curve flattens as  the value of $\eta$ grows.

The density function for the return $Z$ under this model is given by
\begin{equation*}
f_{\mathcal{G}^{H}}( z) =\frac{n^{n}}{\Gamma ( n) }%
\sqrt{\frac{2\omega \eta }{\pi }}e^{\omega }\Bigl( \frac{\omega }{\eta
( \omega \eta +2nz) }\Bigr) ^{n/2+1/4}
z^{n-1}K_{n+1/2}\Bigl( \sqrt{\frac{\omega }{\eta }( \omega \eta
+2nz) }\Bigr) ,
\end{equation*}
with $\omega ,\eta ,z>0$ and $n\geq 1$.
The $r$-th moment of the $\mathcal{G}^{H}$ distribution is
\begin{equation*}
E_{\mathcal{G}^{H}}( Z^{r}) =\left( \frac{\eta }{n}\right)
^{r}e^{\omega }\sqrt{\frac{2\omega }{\pi }}K_{r-1/2}( \omega )
\frac{\Gamma ( n+r) }{\Gamma ( n) },
\label{ec_moment_gh}
\end{equation*}%
which is used for parameter estimation.
The modified Bessel function of the third kind and order $\nu $, whose integral representation is, according to~\cite{Gradshteyn80}, given by:
$$
K_\nu(z) = \int_0^\infty \exp\{-z\cosh(t)\} \cosh(\nu t)dt
$$
is here denoted $K_{\nu }$.
Numerical problems arise when computing this function (c.f. \cite{gordonritcey}), but the distribution used in this paper circumvents this issue as will be seen in the next section.

\appsection{Polarimetric Laws under the Multiplicative Model}\label{app:polarimetric}

This section presents the distributions for polarimetric SAR data: the Complex Multivariate Gaussian distribution and the Centered Wishart distribution (see \cite{Goodman63b,Goodman63a,goodman85,Srivastava65}).
This last one is the most frequently used model for the return coming from homogeneous areas, and serves as a basis for the return from heterogeneous and very heterogeneous areas \cite{oliverquegan98}.

\subsection{Complex Multivariate Gaussian distribution}

Let $\mathbf{T}= [t_1, \dots, t_m]$ be a random $m$-dimensional complex vector with Gaussian Complex Multivariate i.i.d components. Each component has the form $t_k= R_k+i I_k$ where $R_k$ and $I_k$ are the real and imaginary parts, respectively, and they are real random variables.
Then, we can define a $2m$-dimensional Gaussian Multivariate vector $\mathbf{H}=[R_1,I_1, \dots, R_m,I_m]$. Its covariance matrix $\Sigma_{\mathbf{H}}$
is a block matrix with blocks given by $(\Sigma_{\mathbf{H}})_{k,\ell}$ of size $2 \times 2$ with $k,\ell=1,\ldots,m$, as follows
\begin{align*}
(\Sigma_{\mathbf{H}})  _{k,\ell}  & =\mathbb{E}\left[
\begin{array}
[c]{cc}%
(R_k-\mu_{R_k}) (R_{\ell}-\mu_{R_{\ell}})  & (  R_{k}-\mu_{R%
_k})  ( I_{\ell}-\mu_{I_{\ell}}) \\
( I_{k}-\mu_{I_{k}})  (  R_{\ell
}-\mu_{R_\ell})  & (I_{k}-\mu_{I%
_k})  ( I_\ell-\mu_{I_\ell})
\end{array}
\right] \nonumber\\
& = \left\{
\begin{array}
[c]{ccc}%
\frac{\sigma_{k}^{2}}{2}\left[
\begin{array}
[c]{cc}%
1 & 0\\
0 & 1
\end{array}
\right]  & \text{if} & k=\ell\\
\frac{\sigma_{k}\sigma_{\ell}}{2}\left[
\begin{array}
[c]{cc}%
a_{k\ell} & -b_{k\ell}\\
b_{k\ell} & a_{k\ell}
\end{array}
\right]  & \text{if} & k\neq \ell
\end{array}
\right.
\label{MatrBloquesH}%
\end{align*}
where $\sigma_{k}/\sqrt{2}$ are the standard deviation of each component of the random vector $\mathbf{H}$,  $a_{k\ell}$ and $b_{k\ell}$ are the correlation coefficients.

If the random vector $\mathbf{H}$ is multivariate normal distributed, then the corresponding random vector $\mathbf{T}$ follows a Complex Multivariate Gaussian distribution, which denoted by $\mathbf{T}\sim\mathcal{N}_{C}\left(  \mu_{\mathbf{T}}%
,\Sigma_{\mathbf{T}}\right)$. The density function is given by
\[
f_{\mathbf{T}}\left(  T\right)  =\frac{1}{\pi^{m}\left|  \Sigma_{\mathbf{T}%
}\right|  }\exp\left(  -\left(  T-\mu_{\mathbf{T}}\right)  ^{\ast t}%
\Sigma_{\mathbf{T}}^{-1}\left(  T-\mu_{\mathbf{T}}\right)  \right)  ,
\]
where $\mu_{\mathbf{T}}$  is the mean value and the covariance matrix
\begin{equation}
\left(  \Sigma_{\mathbf{T}}\right)  _{k,\ell}=\left\{
\begin{array}
[c]{ccc}%
\sigma_{k}^{2} & \text{if} & k=\ell\\
\left(  a_{k\ell}+jb_{kl}\right)  \sigma_{k}\sigma_{\ell} & \text{if} &
k\neq \ell,
\end{array}
\right.  \label{MatrCovNormComplMultivar}%
\end{equation}
with $k,\ell=1,\dots,m$.

The Multivariate Complex Gaussian distribution is the base of the  Centered
Complex Wishart distribution which is the return model  corresponding to
homogeneous areas.

\subsection{Centered Complex  Wishart Distribution }\label{wishart}

The Centered Complex  Wishart Distribution is used to model the
\emph{speckle} noise of polarimetric data,  the $n$ \emph{looks} are
considered as $n$ random vectors $\mathbf{T}\left(  1\right)
,\dots,\mathbf{T}\left(  n\right)  $, i.i.d, with $\mathbf{T}\left(
k\right) \sim\mathcal{N}_{C}\left(  0,\Sigma_{\mathbf{T}}\right)  $ whose
dimension is $m$, $m \leq n$.

The random matrix $\mathbf{W}$ of $m\times m$ is defined as:
\begin{equation}
\mathbf{W}=\sum_{k=1}^{n}\mathbf{T}(k)  \mathbf{T}(
k)  ^{\ast t}\text{.}\label{FormulaWapartirdeT}%
\end{equation}

Then, the joint distribution of the $m\times m$ elements of the $\mathbf{W}$ matrix is the Centered Complex Wishart distribution \cite{goodman85} and is denoted as $\mathbf{W}\sim\mathcal{W}\left(
\Sigma_{\mathbf{T}},n\right)$, and the parameter $n$ indicates the degrees
of freedom.

The density function of the random matrix $\mathbf{W}$ is given by:
\begin{equation*}
f_{\mathbf{W}}\left(  W\right)  =\frac{\left|  W\right|  ^{n-m}}{\pi^{m(
m-1)/2}\Gamma\left(  n\right)  \cdots\Gamma\left(  n-m+1\right)
\left|  \Sigma_{\mathbf{T}}\right|  ^{n}}\exp\left(  -\mathrm{tr}\left(
\Sigma_{\mathbf{T}}^{-1}W\right)  \right)  ,\label{dens_wishart}%
\end{equation*}
for $n\geq m$ and for all $W\in\mathbb{C}^{m\times m}$.

In polarimetric SAR images analysis this distribution is used to describe
homogeneous areas. The parameters, that characterize each different region
on the image, are the matrix values given by the
equation~(\ref{MatrCovNormComplMultivar}).

In order to estimate the distribution parameters, we use the principal
diagonal components of the $\mathbf{W}$, defined in
equation~(\ref{FormulaWapartirdeT}) and given by:
\begin{equation*}
W_{i,i}=\sum_{i=1}^n\left|T_i(k)\right|^2, \ i\in \{1,\dots,m\}
\end{equation*}
It holds that $W_{i,i}\sim n \sigma_i^2 \Gamma(n,2n)$, with $\sigma_i^2= \mathbb{E}\left[\left|T_i(k)\right|^2\right]$ for all $ k=1,\dots,n$.

If a random variable $\tilde{W}$ has a $\Gamma(n,2n)$ distribution, then
$\mathbb{E}(\tilde{W})=1$, and $\mathbb{E}(W_{i,i})=n\sigma_i^2$ for every $i$.

From the moments method the estimator of $\sigma_i$ results
$$
\hat{\sigma_i}=\sqrt{\frac{m_1W_{i,i}}{n}}.
$$

If the areas have different degrees of heterogeneity, it is necessary to generalize this model introducing the possibility of having  variable characteristics instead constant features.
To this end, Yueh et al.~\cite{KDistributionPolarimetricPIER90} proposed the $\mathcal{K}$ polarimetric distribution.
In the practice, we find extremely heterogeneous data, for this reason a more flexible and tractable distribution was proposed, called the Harmonic polarimetric distribution and denoted as $\mathcal{G}_{P}^{H}$.

\subsection{Drawing outcomes from the $\mathcal{G}^H$ distribution}
\label{app:simulation}

The $\mathcal{G}^H$ distribution belongs to the multiplicative model, and it describes the law that governs the product $\mathbf Z=X\mathbf Y$, where the independent random variables $X$ and $\mathbf Y$ follow the Inverse Gaussian and Complex Wishart distributions, respectively.

Outcomes from the Complex Wishart distribution can be obtained from transformations of Complex Gaussian distributed outcomes, while outcomes from the Inverse Gaussian distribution are obtained by an acceptance-rejection technique.
These procedures are described in the following.

\subsubsection{Multivariate Gaussian Random variable generation \label{genNM}}

If a random vector $\mathbf{Y}=\left[  Y_{1},\dots,Y_{p}\right] ^{t}$ is Multivariate Normal distributed then,  its density function  is given by
\[
f_{\mathbf{Y}}\left(  \mathbf{y}\right)  =\frac{1}{\left(  2\pi\right)  ^{n/2}\left|
\Sigma_{\mathbf{Y}}\right|  ^{1/2}}\exp\left(  -\frac{1}{2}\left(
\mathbf{y}-\mu_{\mathbf{Y}}\right)  ^{t}\Sigma_{\mathbf{Y}}^{-1}\left( \mathbf{y}-\mu
_{\mathbf{Y}}\right)  \right)  ,
\]
where $\mathbf{y}=\left[  y_{1},\dots,y_{p}\right] ^{t} $, $\mu_{\mathbf{Y}}$ is the mean vector of $\mathbf{Y}$, and
$\Sigma_{\mathbf{Y}}$ is the covariance matrix, which is a symmetric positive definite matrix.

The elements $s_{ij}$ of  $\Sigma_{\mathbf{Y}}$ are given by $s_{ij}=\rho_{ij}\sigma_{i}\sigma_{j}$, $1\leq i,j\leq p$, where $\sigma_{i}$ is the standard deviation of the random variable $Y_{i}$, and $\rho_{ij}$ is the correlation coefficient between  $Y_{i}$ and $Y_{j}$.
It is possible to verify that $\left|  \rho_{ij}\right|  \leq1$, that $\rho_{ij}=\rho_{ji}$ and that $\rho_{ii}=1$ for all $1\leq i\leq p$.

Let $\Phi_{\mathbf{Y}}$ be the $p\times p$-dimension matrix whose columns are the normalized eigenvectors of the matrix $\Sigma_{\mathbf{Y}}$ and let $\Lambda_{\mathbf{Y}}$ be the diagonal matrix with the $p$ eigenvalues of the $\Sigma_{\mathbf{Y}}$ in the diagonal elements, then we have $\Sigma_{\mathbf{Y}}\Phi_{\mathbf{Y}}=\Phi_{\mathbf{Y}}\Lambda_{\mathbf{Y}}$.

In order to generate multivariate normal random values  $\mathbf{Y}$ with
mean value $\mu_{\mathbf{Y}}$ and covariance matrix $\Sigma_{\mathbf{Y}}$, a set of decorrelated zero-mean normal values $\mathbf{W}$ ($\Sigma_{\mathbf{W}}=I$ and $\mu_{\mathbf{W}}=0$), are generated.
They are the transformed by $\mathbf{Y}=\Phi_{\mathbf{Y}}\Lambda_{\mathbf{Y}}^{1/2}\mathbf{W}+\mu_{\mathbf{Y}}$.

\subsubsection{Inverse Gaussian distribution generation}

Algorithm~\ref{AlgoGI} shows how to generate samples from the  $IG(\omega, \eta)$ distribution~\cite{Devroye86}.

\begin{algorithm}
\caption{Inverse Gaussian distribution generation.}\label{AlgoGI}
\begin{algorithmic}[1]
\STATE Generate $t$,  sample of the random variable $T\sim\mathcal{N}\left(  0,1\right) $
\STATE Calculate $v=\eta+\frac{\eta t}{2\omega}-\frac{\eta}{2\omega}%
\sqrt{t\left(  4\omega+t\right)  }$
\STATE Generate $u$, sample of the random variable $U \sim\mathcal{U}_{\left(  0,1\right)  }$
 \IF{$u>1/\left(  1+v\eta\right)  $} \STATE  return $\left(  \eta
^{2}u\right)  ^{-1}$
\ELSE \STATE return $v$ \ENDIF
\end{algorithmic}
\end{algorithm}

\bibliographystyle{abbrv}
\bibliography{bib-juliana2}

\end{document}